\documentclass[10pt,twocolumn,letterpaper]{article}

\usepackage{cvpr}              
\usepackage{booktabs}
\usepackage{array}
\usepackage{tabularx}
\usepackage{enumitem}
\usepackage{threeparttable}
\usepackage{multicol}
\usepackage{makecell}
\usepackage{hyphenat}
\usepackage{longtable}
\usepackage[accsupp]{axessibility}
\definecolor{cvprblue}{rgb}{0.21,0.49,0.74}
\usepackage[pagebackref,breaklinks,colorlinks,allcolors=cvprblue]{hyperref}

\def\paperID{*****} 
\def\confName{CVPR}
\def\confYear{2026}

\title{NTIRE 2026 Challenge on Robust AI-Generated Image Detection in the Wild}

\author{Aleksandr Gushchin~\thanks{A.~Gushchin (alexander.gushchin@graphics.cs.msu.ru), K.~Abud, E.~Shumitskaya, A.~Filippov, G.~Bychkov, M. Erofeev, S.~Lavrushkin, M.~Erofeev, A.~Antsiferova, D.~Vatolin, C. Chen, S. Tan and R.~Timofte were the challenge organizers, while the other authors participated in the challenge. \cref{affilations} contains the authors’ teams and affiliations. NTIRE 2026 webpage: \url{https://cvlai.net/ntire/2026/}} \and Khaled Abud \and Ekaterina Shumitskaya \and Artem Filippov \and Georgii Bychkov \and Sergey Lavrushkin \and Mikhail Erofeev \and Anastasia Antsiferova \and Changsheng Chen \and Shunquan Tan \and Radu Timofte \and Dmitry Vatolin \and Chuanbiao Song \and Zijian Yu \and Hao Tan \and Jun Lan \and Zhiqiang Yang \and Yongwei Tang \and Zhiqiang Wu \and Jia Wen Seow \and Hong Vin Koay \and Haodong Ren \and Feng Xu \and Shuai Chen \and Ruiyang Xia \and Qi Zhang \and Yaowen Xu \and Zhaofan Zou \and Hao Sun \and Dagong Lu \and Mufeng Yao \and Xinlei Xu \and Fei Wu \and Fengjun Guo \and Cong Luo \and Hardik Sharma \and Aashish Negi \and Prateek Shaily  \and Jayant Kumar \and Sachin Chaudhary \and Akshay Dudhane \and Praful Hambarde \and Amit Shukla \and Zhilin Tu \and Fengpeng Li \and Jiamin Zhang \and Jianwei Fei \and Kemou Li \and Haiwei Wu \and Bilel Benjdira \and Anas M. Ali \and Wadii Boulila \and Chenfan Qu \and Junchi Li
}

\begin{document}
\maketitle
\begin{abstract}
This paper presents an overview of the NTIRE 2026 Challenge on Robust AI-Generated Image Detection in the Wild, held in conjunction with the NTIRE workshop at CVPR 2026. The goal of this challenge was to develop detection models capable of distinguishing real images from generated ones in realistic scenarios: the images are often transformed (cropped, resized, compressed, blurred) for practical usage, and therefore, the detection models should be robust to such transformations. The challenge is based on a novel dataset consisting of 108,750 real and 185,750 AI-generated images from 42 generators comprising a large variety of open-source and closed-source models of various architectures, augmented with 36 image transformations. Methods were evaluated using ROC AUC on the full test set, including both transformed and untransformed images. A total of 511 participants registered, with 20 teams submitting valid final solutions. This report provides a comprehensive overview of the challenge, describes the proposed solutions, and can be used as a valuable reference for researchers and practitioners in increasing the robustness of the detection models to real-world transformations.
\end{abstract}    
\section{Introduction}
\label{sec:intro}

The emergence of generative AI has made the synthesis of photorealistic images increasingly accessible. Modern generative models — including diffusion-based architectures~\cite{rombach2022high}, generative adversarial networks~\cite{goodfellow2020generative}, and autoregressive image generators~\cite{li2024autoregressive} --- are capable of producing images that are visually indistinguishable from real photographs. The widespread availability of such models through consumer-facing applications has led to an unprecedented volume of AI-generated imagery in circulation, raising serious concerns in the areas of media integrity, misinformation, digital forensics, and content authentication.

In recent years, the research community has developed a variety of methods for detecting AI-generated images. Early approaches~\cite{wang2020cnn} exploited generator-specific artifacts such as spectral periodicities characteristic of GAN upsampling~\cite{frank2020leveraging} and checkerboard patterns introduced by transposed convolutions~\cite{tan2024rethinking}. More recent methods target the residual artifacts of diffusion model denoising pipelines~\cite{wang2023dire}, or leverage large pretrained vision encoders fine-tuned on real/fake corpora~\cite{ojha2023towards}.

Several prior works and competitions have contributed datasets and benchmarks for AI-generated image detection~\cite{zhu2023genimage, hong2025wildfake, yan2024df40}. Despite notable progress, a fundamental and persistent limitation of existing detectors is their lack of robustness to real-world image transformations. In practical deployment scenarios, images are routinely cropped, resized, recompressed, and blurred before they are encountered by a detector --- operations that can substantially degrade detection performance~\cite{li2025bridging}.  However, existing benchmarks typically either cover a limited number of generators or do not systematically account for real-world image transformations, or evaluate both aspects in isolation. Therefore, the creation of modern detectors that are robust to such transformations remains a serious gap in AI-generated image detection.

This paper presents the NTIRE 2026 Challenge on Robust AI-Generated Image Detection in the Wild, which aims to explore and enhance the robustness of detectors to real-world image transformations, including operations that can substantially disrupt detectors trained on clean data. The challenge is built around a novel large-scale dataset comprising 108,750 real images and 185,750 AI-generated images sourced from 42 generators that span a wide range of open-source and closed-source architectures. Crucially, the dataset incorporates 36 image transformation types reflecting realistic post-processing and distribution conditions, enabling a principled assessment of detector robustness.

This paper provides an overview of the methods submitted to the challenge and reports on their performance. The ideas proposed in this challenge can be used to enhance the robustness of the detectors to real-world transformations, in addition to increasing the detection accuracy that is important in critical spheres.

This challenge is one of the challenges associated with the NTIRE 2026 Workshop~\footnote{\url{https://www.cvlai.net/ntire/2026/}} on:
deepfake detection~\cite{ntire26deepfake}, 
high-resolution depth~\cite{ntire26hrdepth},
multi-exposure image fusion~\cite{ntire26raim_fusion}, 
AI flash portrait~\cite{ntire26raim_portrait}, 
professional image quality assessment~\cite{ntire26raim_piqa},
light field super-resolution~\cite{ntire26lightsr},
3D content super-resolution~\cite{ntire263dsr},
bitstream-corrupted video restoration~\cite{ntire26videores},
X-AIGC quality assessment~\cite{ntire26XAIGCqa},
shadow removal~\cite{ntire26shadow},
ambient lighting normalization~\cite{ntire26lightnorm},
controllable Bokeh rendering~\cite{ntire26bokeh},
rip current detection and segmentation~\cite{ntire26ripdetseg},
low light image enhancement~\cite{ntire26llie},
high FPS video frame interpolation~\cite{ntire26highfps},
Night-time dehazing~\cite{ntire26nthaze,ntire26nthaze_rep},
learned ISP with unpaired data~\cite{ntire26isp},
short-form UGC video restoration~\cite{ntire26ugcvideo},
raindrop removal for dual-focused images~\cite{ntire26dual_focus},
image super-resolution (x4)~\cite{ntire26srx4},
photography retouching transfer~\cite{ntire26retouching},
mobile real-word super-resolution~\cite{ntire26rwsr},
remote sensing infrared super-resolution~\cite{ntire26rsirsr},
AI-Generated image detection~\cite{ntire26aigendet},
cross-domain few-shot object detection~\cite{ntire26cdfsod},
financial receipt restoration and reasoning~\cite{ntire26finrec},
real-world face restoration~\cite{ntire26faceres},
reflection removal~\cite{ntire26reflection},
anomaly detection of face enhancement~\cite{ntire26anomalydet},
video saliency prediction~\cite{ntire26videosal},
efficient super-resolution~\cite{ntire26effsr},
3d restoration and reconstruction in adverse conditions~\cite{ntire26realx3d},
image denoising~\cite{ntire26denoising},
blind computational aberration correction~\cite{ntire26aberration},
event-based image deblurring~\cite{ntire26eventblurr},
efficient burst HDR and restoration~\cite{ntire26bursthdr},
low-light enhancement: `twilight cowboy'~\cite{ntire26twilight},
and efficient low light image enhancement~\cite{ntire26effllie}.

\section{Challenge}
Our challenge is dedicated to the task of distinguishing AI-generated images produced in Text-to-Image setting from real imagery. We focus on the most general image domain, without confining to particular subdomains such as faces, humans, or specific object classes. The goals of this competition is threefold: 
1) Assess the current state of AI-powered image generation and its differentiability from the real images; 
2) Evaluate the detection robustness under complex image degradation pipelines; 
3) Appraise the generalization capabilities of the modern detectors to unseen generators.


\subsection{Dataset}
For this challenge, we introduce our novel dataset containing both high-quality real-world imagery and AI-generated images sourced from a diverse set of generators, ranging from older Stable Diffusion~\cite{rombach2022high} models up to the most recent models to date (e.g., Nano Banana 2~\cite{nanobanana2}, SeeDream 5 Lite~\cite{seedream5}). 

\textbf{Real subset.} To construct a diverse and comprehensive set of real images that represent “in-the-wild” content, we collect them from 3 large-scale web-sourced image-text datasets: CC12M~\cite{changpinyo2021conceptual}, CommonPool~\cite{gadre2023datacomp} and RedCaps~\cite{desai2021redcaps}. After initial filtering for explicit and inappropriate content and data availability check, a total of $\sim$12M images were selected. To increase data quality, we employed series of additional filtering stages, ranging from resolution thresholding and CLIP-deduplication to complex VLM-based image categorization and scoring, further reducing the number of suitable images by 90\%. Finally, we sample 100,000 unique real images from our filtered set for the training stage, and $\sim$9k in total for validation and test splits.

\textbf{Generated subset.} To produce generated images, we utilize 42 open-source and proprietary Text-to-Image generators released between 2022 and 2026. Generation prompts for this subset are collected from the corresponding real images: we first employ a Large Vision-Language model to produce detailed image captions and then rewrite them into concise and structured prompts using LLM. By “pairing” generated images with their real counterparts, we ensure that both subsets reflect similar semantics and content distribution, which should help detectors learn content-agnostic features. To further minimize potential biases in generated imagery, we also align its distributions of resolutions, aspect ratios, JPEG compression quality factors, and other statistics to those of the real subset. For the training split, for each real image (out of 100k), we randomly select from 1 to 3 generators to produce corresponding generated images, resulting in $\sim$177k samples. The training split covers 20 open-source generators in total, with the most recent models including Flux Kontext (dev)~\cite{labs2025flux}, DeepFloyd-IF~\cite{deepfloyd}, and Ovis-image~\cite{wang2025ovis_image}. Most of the top-performing open-source models (e.g. Qwen-Image~\cite{wu2025qwen}, HiDream~\cite{hidreami1technicalreport}, etc.) as well as proprietary generators (Nano Banana~\cite{nanobanana}, Grok Imagine~\cite{grok}, etc.) were reserved for validation and test splits, with each consecutive split containing progressively larger share of state-of-the-art models. For validation and test splits, we use only unique images without its paired counterpart to avoid potential advantage from selecting between multiple similar images. More details on the contents of each split can be found in Appendix in Tables 3 and 4.

\textbf{Robust detection track.} Our competition includes a special track designed to evaluate detectors’ performance in a more challenging scenario where input images might be significantly distorted to conceal its nature. This track reflects various image corruptions and artifacts which might be encountered in real-world detector deployment. To this purpose, we employed an image degradation pipeline inspired by \cite{arniqa}. Each image is transformed with 1 to 5 randomly sampled consecutive distortions from different groups (e.g., noise, compression, blur, etc.). Each transformation has multiple magnitude levels, which are sampled independently on each application. In the robust detection track, both real and generated images undergo the same degradation pipeline. To avoid detector tuning for specific distortion types, we employed different sets of distortions in each dataset split and progressively complicated them in the last stages of the challenge. To identify the most challenging distortion types, we utilized our pre-trained detectors and analyzed their performance on degraded images. In total, we employ 36 different transformation types across the challenge stages, ranging from simple transforms like Gaussian blur and white noise, up to complex techniques like invisible image watermarking
~\cite{yakushev2025wibe}
, neural compression~\cite{cheng2020learned, jpeg_ai_standard}
and watermark-erasing adversarial attacks~\cite{soucek2025wmforger}.
Full list of the transformations can be found in Tables 3 and 4. 
In validation and test splits, half of the real and generated images are selected for the robust track and distorted with the aforementioned pipeline.

\subsection{Evaluation metrics}

\begin{itemize}
    \item Primary metric -- Robust ROC AUC, computed over all distorted images in the set. It measures the detector’s global discriminative ability under complex postprocessing pipelines.
    \item Secondary metric -- Clean ROC AUC, computed over all non-distorted images in the set. It measures detector performance in a clear setting, disregarding potential image transformations.
\end{itemize}
ROC AUC is measured between binary labels (0/1), where 1 indicates a fake image, and the confidence scores submitted by the participants. We selected ROC AUC over Accuracy or F1 score, as it does not require quantizing participants' predictions to binary scores with a predefined threshold. 

\section{Challenge Methods and Teams}

\subsection{MICV}\label{MICV}


\subsubsection{Technical Details}\label{technical-details}

\paragraph{Overview}\label{overview}

In this work, we propose a robust ensemble-based framework for
AI-generated image detection, specifically designed to address the
challenges of domain generalization and cross-platform detection. Our
approach centers on three core pillars:

\begin{itemize}
\item
  Hierarchical Data Strategy: We curate a large-scale, multi-source
  training corpus that integrates open-source academic benchmarks,
  synthetic images from cutting-edge generative models, and
  high-fidelity samples from closed-source commercial APIs, ensuring
  comprehensive coverage of diverse generative artifacts.
\item
  Ensemble-based Architecture: We utilize the powerful representation
  capabilities of multiple DINOv3 backbones, organized into two distinct
  model committees. By employing a late-fusion strategy that averages
  the detection probabilities derived from these ensembles, we achieve a
  more holistic and discriminative signature of AI-generated content.
\item 
  Robust Augmentation: To enhance robustness against ``in-the-wild''
  degradations, we implement a hierarchical, difficulty-aware data
  augmentation pipeline. Combined with a Focal Loss-driven optimization
  and Stochastic Weight Averaging (SWA), our approach effectively bridges
  the distribution gap between laboratory benchmarks and real-world
  unconstrained imagery.
\end{itemize}

\paragraph{Data Collection}\label{data-collection}

High-quality, large-scale, and diverse datasets are fundamental to the
robustness and generalizability of AI-Generated Image detection. To
effectively emulate the complex, ``in-the-wild'' distribution of
generative artifacts, we curate a comprehensive training corpus
comprising millions of samples. Our data acquisition strategy is
hierarchical, categorized into four primary tiers:

\begin{itemize}
\item
  Open-Source Datasets: To establish a solid foundation for cross-domain
  generalization, we integrate a diverse array of open-source resources,
  ranging from established academic benchmarks to large-scale
  repositories hosted on platforms like HuggingFace. This collection
  includes, but is not limited to, GenImage, WildFake, AIGIBench,
  CommunityForensics, and So-Fake-Set, along with representative
  datasets sourced from open-source communities. By aggregating these
  heterogeneous data sources, we ensure that our model is exposed to a
  wide spectrum of generation paradigms and realworld artifacts,
  significantly enhancing the scalability and effectiveness of our
  model.
\item
  Open-Source Generative Models: To align with the rapid evolution of
  generative architectures, we construct a substantial synthetic dataset
  using state-of-the-art open-source models. Our pipeline encompasses a
  multifaceted range of generation tasks, including Text-to-Image (T2I),
  Image-to-Image (I2I), image editing and inpainting, leveraging
  representative models such as Qwen-Image, Z-Image, and the FLUX series
  to capture the distinct artifacts inherent in contemporary
  architectures.
\item
  Closed-Source Commercial Models: Given the prevalence of closedsource
  platforms in real-world applications, we supplement our corpus with
  high-fidelity samples obtained via official APIs. By integrating
  outputs from industry-leading engines such as Seedream, Kling,
  GPT-Image, and Nano-banana-pro, we effectively mitigate the
  distribution shift between open-source research models and proprietary
  commercial systems, thereby enhancing the detector's practical
  deployment efficacy.
\item
  Challenge-Specific Datasets: To further expand our set of training
  samples, we utilized the image samples provided by the competition
  organizers.
\end{itemize}

\paragraph{Methodology}\label{methodology}

\begin{figure}[t]
    \centering
    \includegraphics[width=\columnwidth]{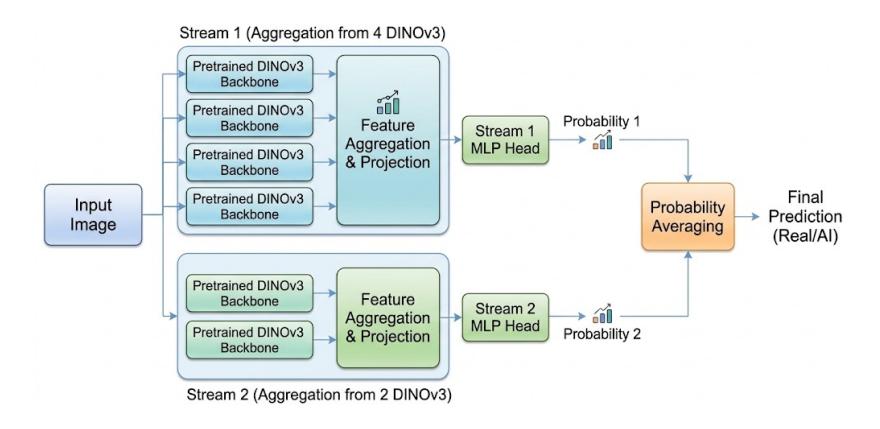}
    \caption{MICV method scheme. DINOv3-based detection framework.}
    \label{MICV_architecture}
\end{figure}

As illustrated in Figure \hyperref[MICV_architecture]{1,} we propose a feature
fusion architecture for AIgenerated image detection. Our approach
leverages the powerful feature representation capabilities of DINOv3
backbones by employing two distinct subnetworks, each designed to
process image features through an ensemble of pretrained backbones.

Specifically, the architecture comprises two independent streams: the
first stream aggregates feature maps from a committee of four DINOv3
backbones, while the second stream integrates features from a separate
committee of two DINOv3 backbones. Within each sub-network, the
aggregated backbone features are processed through a dedicated
projection layer to map them into a latent space, followed by a
multi-layer perceptron (MLP) head that produces the detection
probability. To derive the final prediction, we average output probabilities from both streams.

To bridge the distribution shift between controlled benchmarks and
challenging ``in-the-wild'' imagery, we design a hierarchical,
stochastic data augmentation pipeline structured by difficulty levels.
Our pipeline progresses from simple, individual transformations to
complex combinatorial perturbations. While the former applies individual
degradations such as blur, noise, geometric shifts or compression, the
latter employs multi-stage pipelines that simulate complex degradations.
This hierarchical and multi-faceted augmentation strategy effectively
narrows the domain gap, ensuring superior detection performance even in
highly unconstrained and degraded environments.

\paragraph{Implementation Details}\label{implementation-details}

We initialize our framework using pretrained DINOv3 backbones and
perform end-to-end fine-tuning. During training, images are randomly
cropped and resized to 512 × 512 pixels, supplemented by our
hierarchical augmentation pipeline. During inference, images are
directly resized to 512 × 512 to preserve the global spatial context. The
model is trained on 32 NVIDIA A100 GPUs for 10 epochs, with the entire
procedure completing in approximately 8 hours.

\begin{itemize}
\item
  Objective Function: We employ Focal Loss as our primary objective
  function to address the potential imbalance in sample difficulty and
  to mitigate the dominance of easy-negative samples. The focus
 parameter $\gamma$ is set to 2.0, and the balance parameter $\alpha$ is
  empirically set to 0.5, to ensure that the model focuses on hard-to-classify
  generative artifacts.
\item
  Optimization Strategy: We utilize the AdamW optimizer with a weight
  decay of 0.02. The learning rate is initialized at 1×10-5. To ensure
  training stability, we implement a linear warmup strategy over the
  first epoch, followed by a Cosine Annealing schedule to gradually
  decay the learning rate for the remainder of the training process. We
  employ Stochastic Weight Averaging (SWA) over the final epochs to
  aggregate model weights. This refinement yields a more stable and
  generalized weight configuration, which serves as our final inference
  model.
\item
  Evaluation and Refinement: We evaluate model performance using the
  Area Under the Receiver Operating Characteristic curve (ROC AUC) on a
  dedicated validation set, which is curated by sampling 10,000
  labelbalanced images from the official training corpus. To ensure a
  robust assessment under challenging conditions, we apply a static
  version of the hierarchical data augmentation pipeline to this
  validation set to assess robustness.
\end{itemize}

\paragraph{Acknowledgments}\label{acknowledgments}

We sincerely thank the open-source community, whose contributions have been instrumental to the development of this framework. We are especially grateful to the creators of various AIGC detection benchmarks for providing the foundation of our study, and to the research teams behind state-of-the-art generative models for open-sourcing the architectures and tools that enabled our high-quality data synthesis.

\subsection{Ant International}


\subsubsection{Introduction}
The rapid progress of text-to-image (T2I) generation has made synthetic images increasingly photorealistic, creating a pressing need for robust AI-generated image detection.  
Our solution for the NTIRE 2026 Robust AI-Generated Image Detection Challenge is guided by three principles:
\begin{itemize}
    \item \textbf{Massive and diverse data:} large-scale multi-source training set with extensive augmentations.
    \item \textbf{Model scaling:} leveraging very large vision backbones for improved generalization.
    \item \textbf{Expert ensemble:} combining two complementary specialists trained with different strategies.
\end{itemize}

\begin{figure}[t]
    \centering
    \includegraphics[width=\columnwidth]{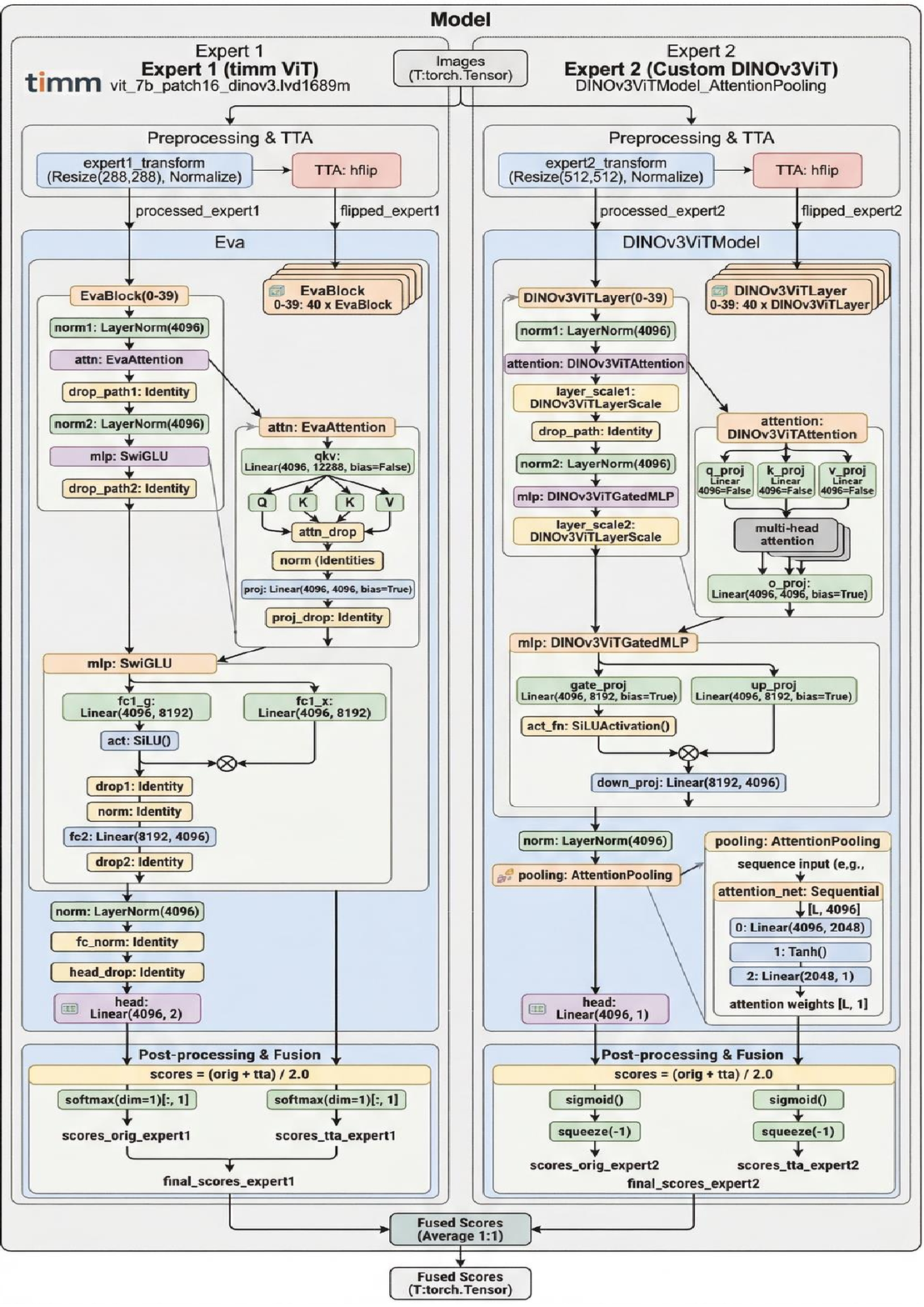}
    \caption{Ant International method scheme.}
    \label{Ant_International_scheme}
\end{figure}

\subsubsection{Data Preparation and Augmentation}
Our training set contains approximately 1 million images from three sources:
\begin{itemize}
    \item Official NTIRE 2026 challenge training set.
    \item Self-generated synthetic images from \textit{LongCat-Image}, \textit{Z-Image}, \textit{Qwen-Image}, and \textit{Qwen-Image-2512}.
    \item Open-source detection datasets: \textit{Dragon} and \textit{UniGenBench}.
\end{itemize}

To improve robustness, we apply a four-level offline augmentation pipeline:
\begin{itemize}
    \item \textbf{Level 1 (Clean):} original images.
    \item \textbf{Level 2 (Mild):} 1--3 random distortions, severity sampled with mean 0 and std 2.5.
    \item \textbf{Level 3 (Moderate):} 3--6 random distortions, mean 2.5 and std 2.0.
    \item \textbf{Level 4 (Heavy):} fixed 6 distortions, mean 3.5, std 1.0.
\end{itemize}

\subsubsection{Backbone Selection}
We evaluated several large-scale vision backbones (\textit{eva02-large}, \textit{eva-giant}, \textit{siglip2-giant}, \textit{DINOv3-7B}).  
Experiments showed strong scaling behavior: larger models and higher input resolutions consistently improved performance and generalization.  
Therefore, we selected \textbf{DINOv3-7B} as the base architecture.

\subsubsection{Dual-Expert Ensemble Architecture}
Our final model is a dual-expert ensemble consisting of two independently fine-tuned \textbf{DINOv3-7B} models (total 14B parameters).  
Given an input image, both experts produce detection scores, which are aggregated into a final robust prediction.  
Both experts are trained with full-parameter fine-tuning on NVIDIA B200 GPUs.

\subsubsection{Expert Training Strategy}

\paragraph{Expert 1: High-Resolution Specialist}
\begin{itemize}
    \item \textbf{Data:} Levels 2, 3, and 4 only.
    \item \textbf{Pooling:} \texttt{attention\_pooling}.
    \item \textbf{Input resolution:} $512 \times 512$.
    \item \textbf{Online augmentation:} random horizontal flip ($p=0.5$) + AugMix (\texttt{m6-w3-d1}).
    \item \textbf{Training setup:} 1 epoch, learning rate $1\times 10^{-6}$, Model EMA, AMP, no dropout.
\end{itemize}

\paragraph{Expert 2: Robustness-Focused Specialist}
\begin{itemize}
    \item \textbf{Data:} Levels 1, 2, 3, and 4.
    \item \textbf{Pooling:} \texttt{first\_token\_pooling}.
    \item \textbf{Input resolution:} $288 \times 288$.
    \item \textbf{Training duration:} 10 epochs.
\end{itemize}


\subsubsection{Inference}
Inference consists of two steps:
\begin{enumerate}
    \item \textbf{Test-time augmentation (TTA):} each expert predicts on multiple augmented variants of each test image.
    \item \textbf{Weighted ensembling:} TTA outputs are aggregated per expert, then combined via weighted averaging across experts.
\end{enumerate}
On a single NVIDIA A100 GPU, the full pipeline runs at approximately \textbf{2.21 images/s} with \textbf{78.25 GB} VRAM usage.

\subsubsection{Conclusion}
Our NTIRE 2026 solution shows that robust AI-generated image detection benefits strongly from:
\begin{itemize}
    \item model scaling (DINOv3-7B),
    \item large and diverse augmented data,
    \item and complementary dual-expert ensembling.
\end{itemize}
By combining a high-resolution specialist and a robustness-focused specialist, we achieve strong generalization under real-world image transformations.

\subsection{TeleAI-TeleGuard. TeleAI Strategy for Robust AI-Generated Image Detection in the Wild}
\label{subsec:teleai_strategy}

\subsubsection{Proposed Method}
\label{subsubsec:proposed_method}

\begin{figure}[t]
    \centering
    \includegraphics[width=\linewidth]{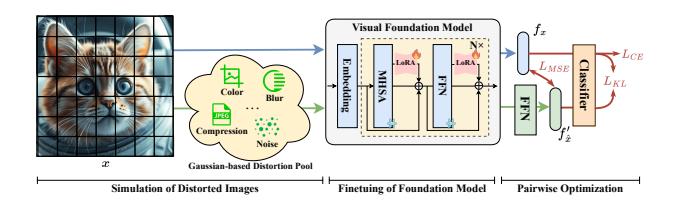}
    \caption{TeleAI-TeleGuard method scheme. \textbf{Pipeline of LPT.} To achieve robust detection, the framework includes distortion simulation, foundation-model fine-tuning, and pairwise optimization.}
    \label{fig:lpt_pipeline}
\end{figure}

We propose a LoRA-based Pairwise Training (LPT) strategy to achieve robust detection of AI-generated images (AIGI). As shown in Figure~\ref{fig:lpt_pipeline}, this strategy consists of three components: foundation-model fine-tuning, distortion simulation, and pairwise optimization.

In addition to the distortions provided by the organizer, after inspecting the validation set we introduce three extra distortion types---\emph{Speckle Noise}, \emph{Color Cast}, and \emph{Organic Moire}---which better match the target data distribution. We also increase distortion severity by setting the mean of the Gaussian distribution to 3.

The foundation model in LPT is EVA-CLIP~\cite{sun2023evaclip}. Following LoRA-style fine-tuning~\cite{hu2022lora}, we adapt the linear layers in the multi-head self-attention (MHSA) and feed-forward network (FFN) of each visual transformer block.

To improve robustness without degrading clean-sample performance, we adopt pairwise optimization by jointly feeding clean images and their corresponding distorted versions in each batch. Features extracted from distorted images are corrected by an additional feed-forward network. The overall training loss is:
\begin{equation}
\mathcal{L} =
\mathcal{L}_{CE}(\mathbf{x}, \mathbf{y})
+ \alpha \cdot \mathcal{L}_{KL}(\mathbf{x}, \hat{\mathbf{x}})
+ \beta \cdot \mathcal{L}_{MSE}(\mathbf{f}_{x}, \mathbf{f}'_{\hat{x}}),
\label{eq:lpt_loss}
\end{equation}
where $\mathbf{x}$, $\hat{\mathbf{x}}$, $\mathbf{f}_{x}$, and $\mathbf{f}'_{\hat{x}}$ denote clean samples, distorted samples, clean-sample features, and corrected distorted-sample features, respectively. $\mathcal{L}_{CE}$, $\mathcal{L}_{KL}$, and $\mathcal{L}_{MSE}$ are cross-entropy, KL divergence, and mean squared error losses. We set $\alpha=0.5$ and $\beta=0.25$.

Experiments are conducted on eight NVIDIA A800 GPUs for 5 epochs. We use AdamW~\cite{loshchilov2017decoupled} with an initial learning rate of $2\times10^{-4}$ and a cosine annealing scheduler~\cite{loshchilov2016sgdr}. To further improve generalization, we additionally include the So-Fake~\cite{huang2025sofake} and Chameleon~\cite{yan2024sanity} datasets during training.

\subsection{INTSIG}
\label{subsec:INTSIG}

\begin{figure}[t]
    \centering
    \includegraphics[width=\columnwidth]{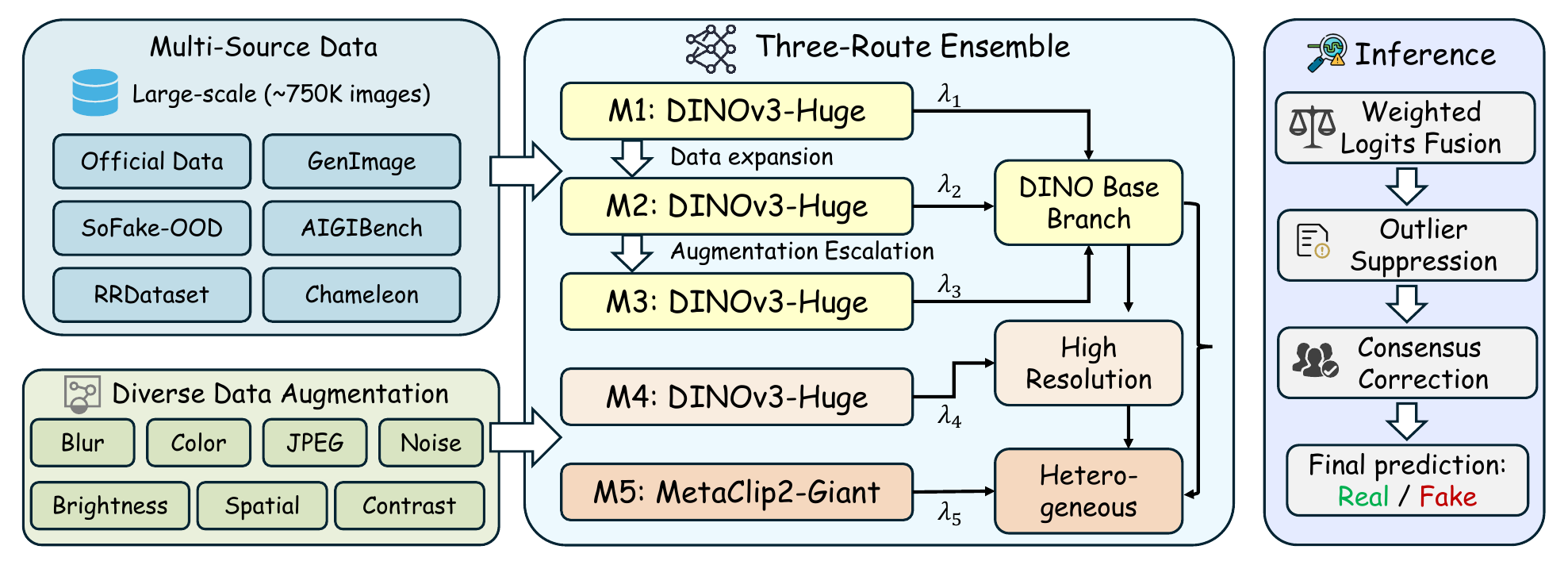}
    \caption{INTSIG method scheme.}
    \label{INTSIG_scheme}
\end{figure}

\subsubsection{Model Training Strategies}
\label{subsubsec:model_training_strategies}

\paragraph{Overview.}
We develop five complementary detectors in a staged pipeline. Model~1 is a full fine-tuning baseline on DINOv3-Huge. Model~2 continues training with expanded data. Model~3 further increases robustness with stronger distortion augmentation. Model~4 introduces a high-resolution branch ($448\times448$). Model~5 replaces the backbone with MetaCLIP2 Giant and applies partial fine-tuning. This design balances in-domain accuracy and OOD robustness.

\paragraph{Model 1: Baseline.}
Model~1 uses DINOv3-Huge with an MLP head:
\begin{equation}
1280 \rightarrow 256 \xrightarrow{\mathrm{ReLU}+\mathrm{Dropout}(0.1)} 2.
\end{equation}
We use AdamW (betas $(0.9,0.999)$) with separate parameter groups: backbone lr $=2\times10^{-5}$, weight decay $=0.05$; head lr $=5\times10^{-4}$, weight decay $=0.01$. The learning-rate schedule is 10\% warmup followed by cosine annealing, stepped per batch. Cross-entropy is used as the training objective, and class imbalance is handled by a distributed weighted sampler.  
Training data includes the official training set, SoFake-OOD, and RRDataset (10\% validation split). Inputs are preprocessed with random resized crop ($256\times256$, bicubic, scale $(0.08,1.0)$, ratio $(0.75,1.3333)$), random horizontal flip, and color jitter. Distortion augmentation is enabled with probability 0.5 (up to 3 distortions, 3 levels). Training runs for 20 epochs on 8$\times$H800 (DDP), with per-GPU batch size 16/8 (train/val).

\paragraph{Model 2: Incremental training with dataset expansion.}
Starting from Model~1, we reduce learning rates (backbone $5\times10^{-6}$, head $5\times10^{-5}$) and expand training data to: official set + SoFake-OOD + RRDataset + Chameleon + 50\% GenImage\_val + AIGIBench\_test. Other settings are unchanged.

\paragraph{Model 3: Incremental training with enhanced augmentation.}
Model~3 continues from Model~2 while increasing distortion intensity to 5 distortions and 5 levels (augmentation probability remains 0.5). Other settings are unchanged.

\paragraph{Model 4: High-resolution path.}
Model~4 uses $448\times448$ inputs to capture finer artifacts. AdamW is retained with 5\% warmup and cosine annealing. Learning rates are set to $5\times10^{-5}$ (backbone) and $1\times10^{-3}$ (head). Training data is the official set only, with a 1\% validation split. Preprocessing follows the same policy as Model~1 but at $448\times448$ resolution. Distortion augmentation uses probability 0.5, with up to 3 distortions and 5 levels.

\paragraph{Model 5: Backbone replacement.}
Model~5 adopts MetaCLIP2 Giant. We partially fine-tune all LayerNorm parameters and the full parameters of the last two layers. 
The classifier is:
\begin{align}
&\text{MLP1: } 1664 \rightarrow 2048 \rightarrow 512, \nonumber\\
&\text{MLP2: } 512 \rightarrow 256 \rightarrow 2.
\end{align}
Training uses the official set, learning rate $10^{-4}$, and FocalLoss. Inputs are resized to $378\times378$; training includes random flip, random rotation, and random augmentation. Distortion augmentation is lighter (probability 0.2, 3 distortions, 3 levels).

\begin{table}[t]
\centering
\small
\setlength{\tabcolsep}{4pt}
\renewcommand{\arraystretch}{1}
\begin{tabular}{p{0.1\columnwidth}p{0.1\columnwidth}p{0.5\columnwidth}p{0.13\columnwidth}}
\hline
Model & Input & Core change & Weight \\
\hline
M1 & $256^2$ & DINOv3-Huge baseline & 0.3675 \\
M2 & $256^2$ & Incremental data expansion & 0.0735 \\
M3 & $256^2$ & \makecell[tl]{Stronger distortion\\augmentation} & 0.0490 \\
M4 & $448^2$ & High-resolution branch & 0.2100 \\
M5 & $378^2$ & MetaCLIP2 Giant backbone & 0.3000 \\
\hline
\end{tabular}
\caption{Summary of training paths and ensemble weights.}
\label{tab:model_paths}
\end{table}

\subsubsection{Model Ensemble Strategy}
\label{subsubsec:ensemble_strategy}

\paragraph{Weighted hierarchical fusion.}
The final logit is computed as:
\begin{align}
\mathrm{Final\_Logit}
&= 0.7\Big[0.7(0.75M1 + 0.15M2 + 0.1M3) \nonumber\\
&\quad + 0.3M4\Big] + 0.3M5 .
\end{align}
Horizontal-flip TTA is applied to M3 and M4.

\paragraph{Dual-gating correction.}
We define directional confidence:
\begin{equation}
\mathrm{diff}=\mathrm{logit}_1-\mathrm{logit}_0.
\end{equation}
\textit{Gate-1 (strong consensus correction)} is triggered when M4 and M5 strongly agree ($|M4_{\mathrm{diff}}|\ge 8$, $|M5_{\mathrm{diff}}|\ge 3$), but the fused output has the opposite direction. We then shift the final logit by 2.5 toward the M4/M5 direction.  
\textit{Gate-2 (anomaly suppression)} is triggered when at least 3 models in $\{M1,M2,M3,M5\}$ agree while M4 disagrees; M4 is excluded and remaining weights are renormalized.

\subsubsection{Engineering Optimization}
\label{subsubsec:engineering_optimization}

\paragraph{Parallel inference.}
Inference runs with one CUDA stream per model, CPU-side multi-threaded preprocessing, and AMP for M1--M4.

\paragraph{I/O and overlap.}
We use asynchronous data loading (\texttt{num\_workers=4}) and prefetching (\texttt{prefetch=4}), overlapping CPU preprocessing with GPU inference to reduce idle time.

\begin{table}[t]
\centering
\small
\setlength{\tabcolsep}{4pt}
\renewcommand{\arraystretch}{1}
\begin{tabular}{p{0.20\columnwidth}p{0.37\columnwidth}p{0.31\columnwidth}}
\hline
Dimension & Technical solution & Effect \\
\hline
\makecell[tl]{GPU\\parallelism} & 5 independent CUDA streams & Full model overlap \\
\makecell[tl]{Compute\\precision} & AMP for M1--M4 & $\sim$30\% throughput gain \\
Data loading & Async I/O + prefetch=4 & Reduced wait time \\
Pipeline & CPU/GPU overlap & Higher utilization \\
\hline
\end{tabular}
\caption{Inference-time engineering optimizations.}
\label{tab:eng_opt}
\end{table}

\subsection{Vincentlc. Robust AI-Generated Image Detection via SigLIP2-Giant and Perturbation-Aware Training}
\label{sec:method_robust_siglip2}

\begin{figure}[t]
    \centering
    \includegraphics[width=\columnwidth]{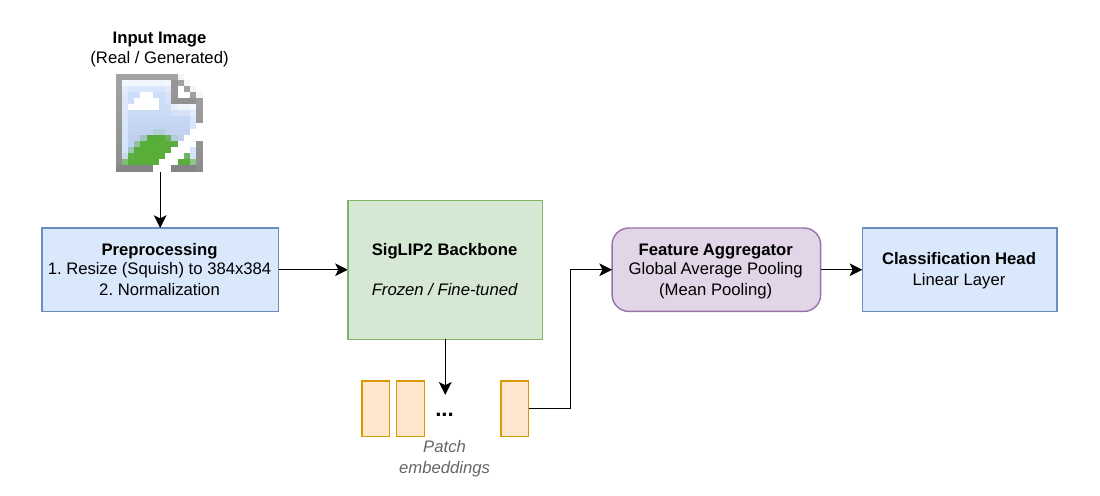}
    \caption{Vincentlc method scheme.}
    \label{Vincentic_architecture}
\end{figure}

\begin{table*}[t!]
\begin{center}
\begin{tabular}{ccccccc}
\hline
& \multicolumn{2}{c}{Open Test} & \multicolumn{2}{c}{Hidden Test} & \multicolumn{2}{c}{Average} \\
\textbf{Method} & ROC AUC & Rob. ROC AUC & ROC AUC & Rob. ROC AUC & ROC AUC & Rob. ROC AUC \\
\hline
MICV & \textbf{0.9978} & \textbf{0.9738} & \underline{0.9969} & \underline{0.9707} & \textbf{0.9974} & \textbf{0.9723} \\
Ant International & \underline{0.9973} & \underline{0.9731} & \textbf{0.9971} & \textbf{0.9711} & \underline{0.9972} & \underline{0.9721} \\
TeleAI-TeleGuard & 0.9762 & 0.9215 & 0.9809 & 0.9286 & 0.9786 & 0.9251 \\
INTSIG & 0.9810 & 0.9090 & 0.9895 & 0.9171 & 0.9897	& 0.9130 \\
vincentlc & 0.9497 & 0.8633 & 0.9557 & 0.8828 & 0.9527 & 0.8730 \\
UESTC & 0.9693 & 0.8558 & 0.9764 & 0.8800 & 0.9729	& 0.8679 \\
Reagvis Labs & 0.9423 & 0.8474 & 0.9481 & 0.8733 & 0.9452	& 0.8603 \\
PSU & 0.9132 & 0.8334 & 0.9322 & 0.8483 & 0.9227	& 0.8408 \\
Shallow Real & 0.9954 & 0.8302 & 0.9951 & 0.8370 & 0.9953	& 0.8336 \\
\end{tabular}
\caption{Final results and rankings of the top-9 teams. Performance is reported on the open and hidden test sets using both clean ROC-AUC and Robust ROC-AUC. The final ranking is determined by the average Robust ROC-AUC across both splits.}
\vspace{-20pt}
\label{tab:final}
\end{center}
\end{table*}

\subsubsection{Overview}
We propose a simple yet effective framework for detecting AI-generated images in the wild.
Our method combines a pre-trained vision--language backbone with a lightweight linear classification head.
To improve robustness under severe real-world distribution shifts, we explicitly integrate the official image distortion pipeline used in challenging evaluation protocols~\cite{li2025bridging} into training, forcing the network to learn perturbation-invariant representations.

\subsubsection{Model Architecture}
\paragraph{Backbone Feature Extractor.}
We experimented with multiple scales of the SigLIP2 family and found that the largest variant, \texttt{siglip2-giant-opt-patch16-384}, achieves the strongest performance.
For downstream binary classification, we extract the patch-token sequence from the final hidden layer of the pre-trained SigLIP2 encoder~\cite{siglip2}.

\paragraph{Feature Aggregation and Classification Head.}
We evaluated several aggregation strategies, including CLS-token extraction, attention pooling, and multi-layer feature concatenation.
Empirically, global average pooling over all final-layer patch tokens yields the most robust and stable results.
The pooled feature vector is then fed into a single linear layer to produce real/fake logits.

\subsubsection{Robust Data Augmentation}
\paragraph{Spatial Transforms (``Squish'' Strategy).}
Instead of random resized cropping (which may remove localized forensic cues), we apply a ``squish'' strategy:
all images are directly resized to $384\times384$ (ignoring aspect ratio), followed by random horizontal flipping.
This preserves full image content while maintaining training efficiency.

\paragraph{Competition-Specific Distortions.}
We wrap the official \texttt{distort\_images} function into the PyTorch transform pipeline.
In our best run, we set \texttt{distortion\_prob=1.0}, so every training image is distorted, with up to three random post-processing operations sampled at multiple severity levels (\texttt{num\_levels=5}).
This aggressive augmentation acts as a strong regularizer and is a key driver of our Robust ROC AUC improvements.

\section{Results}

Table~\ref{tab:final} presents the final scores and rankings of the top-9 participating teams. All submissions were verified by re-running the provided models in the official evaluation environment to ensure consistency with the public leaderboard. 

The final ranking is determined based on the average Robust ROC-AUC across the open and hidden test datasets. For completeness, we also report the average clean ROC-AUC, as well as all individual metrics, including clean and robust ROC-AUC on both dataset splits.

Overall, the top-performing methods achieve ROC AUC values exceeding 0.99. The \textbf{MICV} team ranks first according to the primary metric, achieving the highest average Robust ROC-AUC of 0.9723. Their method also delivers the best average clean ROC-AUC (0.9974), indicating strong performance under both standard and robustness-aware evaluations.
\textbf{Ant International} closely follows in second place, achieving the best performance on the hidden test set for both clean and robust ROC-AUC (0.9971 and 0.9711, respectively). This demonstrates excellent generalization to unseen data. The small margin between the top-2 teams suggests that both approaches are competitive and approach the upper bound of performance on this benchmark.
A clear performance gap is observed between the top-2 teams and the remaining participants. Methods such as \textbf{TeleAI-TeleGuard} and \textbf{INTSIG} achieve strong clean ROC AUC scores (around 0.98–0.99), but show noticeably lower robustness, with Robust ROC-AUC dropping to approximately 0.91–0.93. This indicates increased sensitivity to challenging conditions and highlights robustness as a key differentiating factor. The lower-ranked methods, including \textbf{vincentlc} and \textbf{UESTC}, exhibit further degradation in both clean and robust metrics, particularly in the robustness setting, where performance falls below 0.88. 

\section{Conclusion}
\label{conclusion}

In the NTIRE 2026 Robust AI-Generated Image Detection in the Wild, 500+ registeres participants competed to develop state-of-the-art deepfake detection methods for a novel dataset of 250,000+ images. Most final solutions used Expert-base architecture with several different models and use Transformer-based architectures. The results suggest that, although performance has improved substantially, the problem is not yet solved, and there remains room for advances in design, training strategies and data curation.


\section{Acknowledgments}
The work of Aleksandr Gushchin, Khaled Abud, Ekaterina Shumitskaya, Georgii Bychkov, Sergey Lavrushkin, Anastasia Antsiferova, and Dmitry Vatolin was financially supported by the Institute for Artificial Intelligence of Lomonosov Moscow State University.
The evaluation was carried out using the MSU-270 supercomputer of the Lomonosov Moscow State University. 
Future subjective evaluations of artifact visibility will be conducted using Yandex.Tasks.
The submission system was partially supported by the Humboldt Foundation, OPPO, Kuaishou, and the University of Wurzburg (Computer Vision Lab).

\section{Teams and Affiliations}
\label{affilations}

\subsection{NTIRE 2026 Challenge Organizing team}

\noindent\textbf{\textit{Members:}}\\
Aleksandr Gushchin$^{1}$ (alexander.gushchin@graphics.cs.msu.ru),\\
Khaled Abud$^{1}$ (khaled.abud@graphics.cs.msu.ru),\\
Ekaterina Shumitskaya$^{1}$ (ekaterina.shumitskaya@graphics.cs.msu.ru),\\
Artem Filippov$^{2}$ (artyom.filippov@graphics.cs.msu.ru),\\
Georgii Bychkov$^{1}$ (georgy.bychkov@graphics.cs.msu.ru),\\
Sergey Lavrushkin$^{1}$ (sergey.lavrushkin@graphics.cs.msu.ru),\\
Mikhail Erofeev$^{5}$ (mikhail@erofeev.pw),\\
Anastasia Antsiferova$^{1}$ (Aantsiferova@graphics.cs.msu.ru),\\
Changsheng Chen$^{3}$ (cschen@smbu.edu.cn),\\
Shunquan Tan$^{3}$ (tansq@smbu.edu.cn),\\
Radu Timofte$^{4}$ (radu.timofte@uni-wuerzburg.de),\\
Dmitry Vatolin$^{1}$ (dmitriy@graphics.cs.msu.ru)\\
\noindent\textbf{\textit{Affiliations:}}\\
$^{1}$ MSU Institute for Artificial Intelligence\\
$^{2}$ Lomonosov Moscow State University\\
$^{3}$ Shenzhen MSU-BIT University\\
$^{4}$ University of Würzburg\\
$^{5}$ Independent Researcher\\

\subsection{MICV}
\noindent\textbf{\textit{Members:}}\\
Chuanbiao Song$^{1}$ (songchuanbiao.scb@antgroup.com),\\
Zijian Yu$^{1}$ (yuzijian.yzj@antgroup.com),\\
Hao Tan$^{1}$ (tanhao2023@ia.ac.cn),\\
Jun Lan$^{1}$ (yelan.lj@antgroup.com)\\
\noindent\textbf{\textit{Affiliations:}}\\
$^{1}$ Ant Group, China

\subsection{Ant International}
\noindent\textbf{\textit{Members:}}\\
Zhiqiang Yang$^{1}$ (874125760@qq.com),\\
Yongwei Tang$^{1}$ (1598521844@qq.com),\\
Zhiqiang Wu$^{1}$ (1753940858@qq.com),\\
Jia Wen Seow$^{1}$ (sscarlettss0111@gmail.com),\\
Hong Vin Koay$^{1}$ (koayhv@gmail.com),\\
Haodong Ren$^{1}$ (renhaodong.rhd@ant-intl.com),\\
Feng Xu$^{1}$ (fuyu.xf@ant-intl.com),\\
Shuai Chen$^{1}$ (shuai.cs@ant-intl.com)\\
\noindent\textbf{\textit{Affiliations:}}\\
$^{1}$ Ant International, Singapore

\subsection{TeleAI-TeleGuard}
\noindent\textbf{\textit{Members:}}\\
Ruiyang Xia$^{1}$ (ryon@stu.xidian.edu.cn),\\
Qi Zhang$^{1}$ (zhangq139@chinatelecom.cn),\\
Yaowen Xu$^{1}$ (xuyw1@chinatelecom.cn),\\
Zhaofan Zou$^{1}$ (zouzhf41@chinatelecom.cn),\\
Hao Sun$^{1}$ (sunh10@chinatelecom.cn)\\
\noindent\textbf{\textit{Affiliations:}}\\
$^{1}$ TeleAI-TeleGuard, China

\subsection{INTSIG} \label{INTSIG}
\noindent\textit{\textbf{Members:}}
Fei Wu$^{1,2}$ (wu\_fei@sjtu.edu.cn), \\
Dagong Lu$^{1}$ (dagong\_lu@intsig.net), \\
Mufeng Yao$^{1}$ (mufeng\_yao@intsig.net), \\
Xinlei Xu$^{1}$ (xinlei\_xu@intsig.net), \\
Fengjun Guo$^{1}$ (fengjun\_guo@intsig.net) \\
\noindent\textit{\textbf{Affiliations:}}\\
$^1$ IntSig Information Co. Ltd, Shanghai, China \\
$^2$ Shanghai Jiao Tong University, Shanghai, China

\subsection{Vincentlc}
\noindent\textbf{\textit{Members:}}\\
Cong Luo (1486891983@qq.com)\\

\subsection{Reagvis Labs}
\noindent\textbf{\textit{Members:}}\\
Hardik Sharma$^{1}$ (d25092@students.iitmandi.ac.in),\\
Aashish Negi$^{1}$ (dd25011@students.iitmandi.ac.in),\\
Prateek Shaily$^{1}$ (prateekshaily820@gmail.com),\\
Jayant Kumar$^{1}$ (aryanjay2k05@gmail.com),\\
Sachin Chaudhary$^{1}$ (sachin.chaudhary@ddn.upes.ac.in),\\
Akshay Dudhane$^{1}$ (akshay.dudhane@mbzuai.ac.ae),\\
Praful Hambarde$^{1}$ (praful@iitmandi.ac.in),\\
Amit Shukla$^{1}$ (amitshukla@iitmandi.ac.in)\\
\noindent\textbf{\textit{Affiliations:}}\\
$^{1}$ IIT Mandi, Reagvis Labs, Mandi, India

\subsection{UESTC}
\noindent\textbf{\textit{Members:}}\\
Zhilin Tu$^{1}$ (202521080234@std.uestc.edu.cn),\\
Fengpeng Li$^{1}$ (fengpeng.li@connect.umac.mo),\\
Jiamin Zhang$^{1}$ (18482160071@163.com),\\
Jianwei Fei$^{1}$ (fei\_jianwei@163.com),\\
Kemou Li$^{1}$ (yc47912@umac.mo),\\
Haiwei Wu$^{1}$ (haiweiwu@uestc.edu.cn)\\
\noindent\textbf{\textit{Affiliations:}}\\
$^{1}$ University of Electronic Science and Technology of China, China

\subsection{PSU}
\noindent\textbf{\textit{Members:}}\\
Bilel Benjdira$^{1}$ (bbenjdira@psu.edu.sa),\\
Anas M. Ali$^{1}$ (aaboessa@psu.edu.sa),\\
Wadii Boulila$^{1}$ (wboulila@psu.edu.sa)\\
\noindent\textbf{\textit{Affiliations:}}\\
$^{1}$ Prince Sultan University, Robotics and Internet-of-Things Laboratory, Riyadh, Saudi Arabia

\subsection{ShallowReal}
\noindent\textbf{\textit{Members:}}\\
Chenfan Qu$^{1}$ (202221012612@mail.scut.edu.cn),\\
Junchi Li$^{2}$ (hongge568@126.com)\\
\noindent\textbf{\textit{Affiliations:}}\\
$^{1}$ South China University of Technology, Guangzhou, China\\
$^{2}$ Zhejiang University, Hangzhou, China

\section{Challenge Methods and Teams (Continued)}

\subsection{Reagvis Labs. RAPID: Robust AI-Generated Prototype Image Detection}
\label{rapid-robust-ai-generated-prototype-image-detection}

We propose RAPID, a cascaded six-model ensemble for robust AI-generated image detection under realistic degradations. The system combines four complementary representation families: CLIP~\cite{radford2021clip}, SigLIP~\cite{zhai2023siglip}, SRM-style forensic residual modeling~\cite{fridrich2012rich}, and EVA-02 masked-image pretraining~\cite{fang2024eva02}. Each branch is fine-tuned with parameter-efficient adaptation (mainly LoRA~\cite{hu2022lora}), and final prediction is obtained by staged logit-space fusion. To keep memory low, models are executed sequentially; peak VRAM remains below 4\,GB.

\subsubsection{Architecture}
\label{architecture}

RAPID uses six detectors in a progressive refinement cascade. The main semantic branch (G4) is CLIP ViT-L/14 with LoRA on attention and MLP projections, followed by a GAPL prototype-attention head that compares normalized embeddings against learned forensic prototypes. A high-resolution SigLIP-v2 branch (384$\times$384) shares the same GAPL idea and improves sensitivity to fine local artifacts. In parallel, an SRM+Bayar ForensicCNN branch processes raw pixel inputs without normalization, explicitly modeling residual-domain inconsistencies. Two EVA-02 checkpoints provide complementary MIM-based cues: the original LoRA model and a retrained fixed-initialization variant with improved calibration under degradation. Finally, G4\_v2 (continued training from G4 EMA) performs last-step correction on difficult modern generations.

\begin{figure}[t]
    \centering
    \includegraphics[width=\columnwidth]{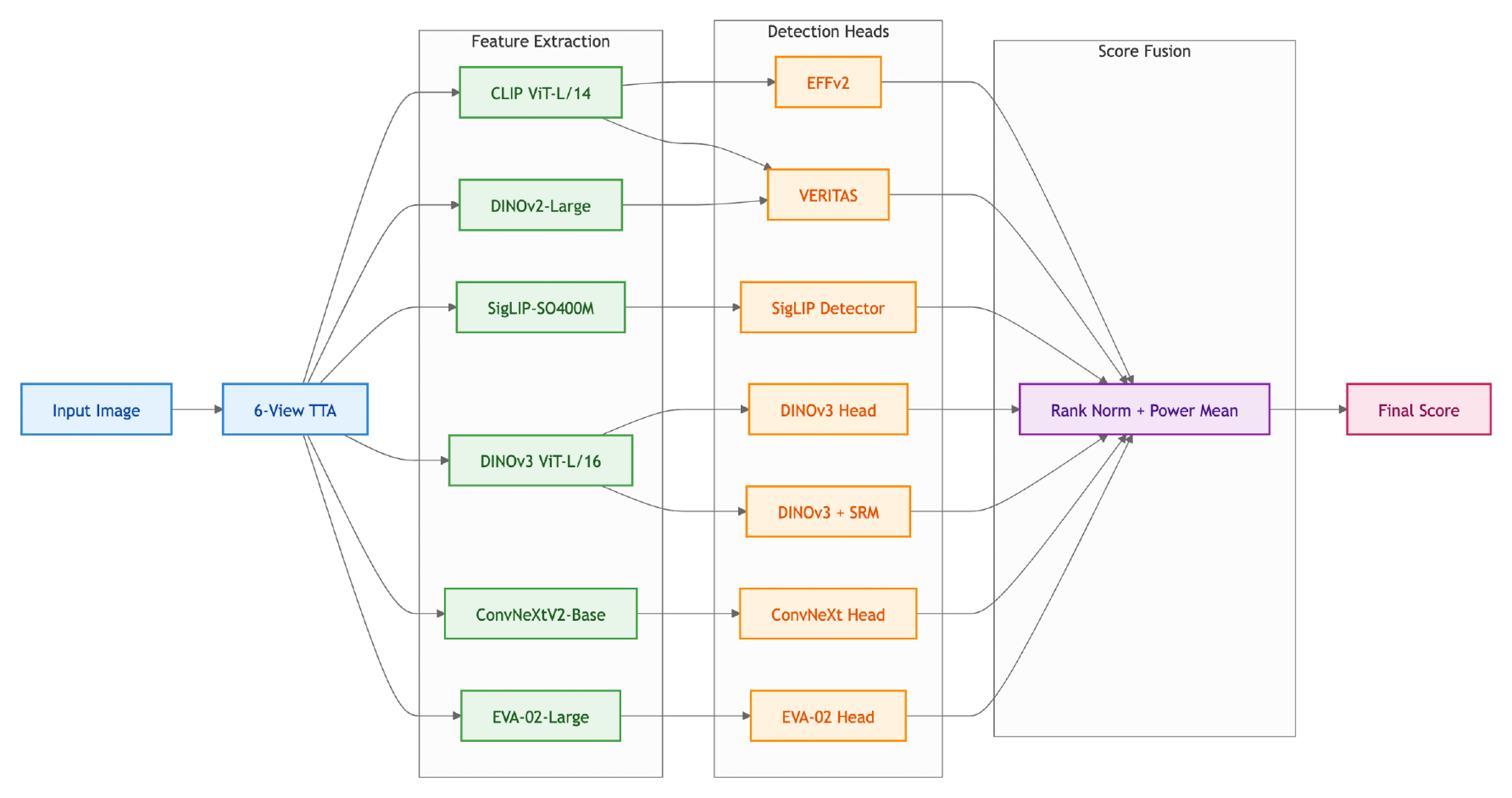}
    \caption{Reagvis Labs method scheme. Overview of RAPID with staged logit fusion.}
    \label{fig:rapid_arch}
\end{figure}

\subsubsection{Logit-Space Cascade Fusion}
\label{score-fusion-4-stage-logit-blend}

All branches output probabilities in $[0,1]$. We combine them in logit space, where additive mixing is better calibrated than direct probability averaging.
\begin{equation}
\ell(p)=\log\frac{p}{1-p}.
\end{equation}

\begin{equation}
\begin{aligned}
s_{1a} &= \sigma\!\left(0.50L_{G4}+0.35L_{S}+0.15L_{R}\right),\\
b      &= \sigma\!\left(0.80L_{1a}+0.20L_{E}\right),\\
s_{2}  &= \sigma\!\left(0.85L_{b}+0.15L_{E_f}\right),\\
f      &= \sigma\!\left(0.89L_{2}+0.11L_{G4v2}\right).
\end{aligned}
\label{eq:rapid_fusion}
\end{equation}

\noindent
Here, $L_{G4}=\ell(G4)$, $L_S=\ell(\text{SigLIP})$, $L_R=\ell(\text{SRM})$, $L_E=\ell(\text{EVA02})$, $L_{E_f}=\ell(\text{EVA02}_{\text{fixed}})$, $L_{1a}=\ell(s_{1a})$, $L_b=\ell(b)$, $L_2=\ell(s_2)$, and $L_{G4v2}=\ell(G4_{\text{v2}})$.

This cascade was built incrementally through validation ablations; each stage was retained only if it improved robust AUC.

\subsubsection{Inference-Time Robustness}
\label{test-time-augmentation}

We apply 8-view TTA per model, mixing geometric and degradation-aware transforms (flip, multi-scale center crops, corner crops, blur, and JPEG perturbation). For sigmoid heads, we average logits across views and then apply sigmoid; for softmax heads, we average class probabilities. SigLIP uses resolution-matched 384px variants of the same views, while SRM uses bicubic resizing on raw $[0,1]$ inputs.

\subsubsection{Training Protocol}
\label{training-details}

All branches are trained on the official NTIRE 2026 data using AdamW with cosine decay and warmup, EMA checkpointing, and mixed precision where stable. Augmentations include random crop/flip, JPEG recompression, Gaussian blur/noise, and random resizing. Transformer branches use LoRA (typically $r=32$), while G4\_v2 is obtained by low-LR continuation from the G4 EMA checkpoint with SAM and CutMix for harder-sample generalization.

\subsubsection{Runtime and Memory}
\label{runtime-performance}

Table~\ref{tab:rapid_runtime} reports per-model latency and memory on an H100.

\begin{table}[t]
\centering
\setlength{\tabcolsep}{3.5pt}
\renewcommand{\arraystretch}{1.05}
\resizebox{\columnwidth}{!}{
\begin{tabular}{lccc}
\toprule
Model & Time (ms) & Peak VRAM (GB) & Fine-tuning \\
\midrule
G4 (CLIP+GAPL)       & $\sim$37 & 2.6 & LoRA $r{=}32$ (attn+MLP) \\
SigLIP-v2            & $\sim$26 & 1.8 & LoRA $r{=}32$ (384px) \\
SRM ForensicCNN      & $\sim$5  & 0.1 & full training (FP32) \\
EVA-02 (original)    & $\sim$25 & 1.3 & LoRA $r{=}32$ + SwiGLU \\
EVA-02 (fixed)       & $\sim$25 & 1.3 & LoRA $r{=}32$ retrain \\
G4\_v2 (CLIP+GAPL)   & $\sim$37 & 2.6 & LoRA continuation \\
\bottomrule
\end{tabular}
}
\caption{Per-branch inference cost and memory. Sequential loading keeps total peak memory below 4\,GB.}
\label{tab:rapid_runtime}
\end{table}

\subsubsection{Design Rationale}
\label{key-design-decisions}

RAPID is built around complementarity: semantic and language-aligned features (CLIP/SigLIP), residual forensic cues (SRM), and MIM-driven representations (EVA-02) fail on different samples, making staged logit blending consistently stronger than any single branch. Prototype-attention heads improve interpretability by comparing each image to learned real/fake forensic anchors, while degradation-aware TTA stabilizes predictions under compression, blur, and scale shifts commonly observed in the wild.

\subsection{UESTC}

\begin{figure}[t]
    \centering
    \includegraphics[width=\columnwidth]{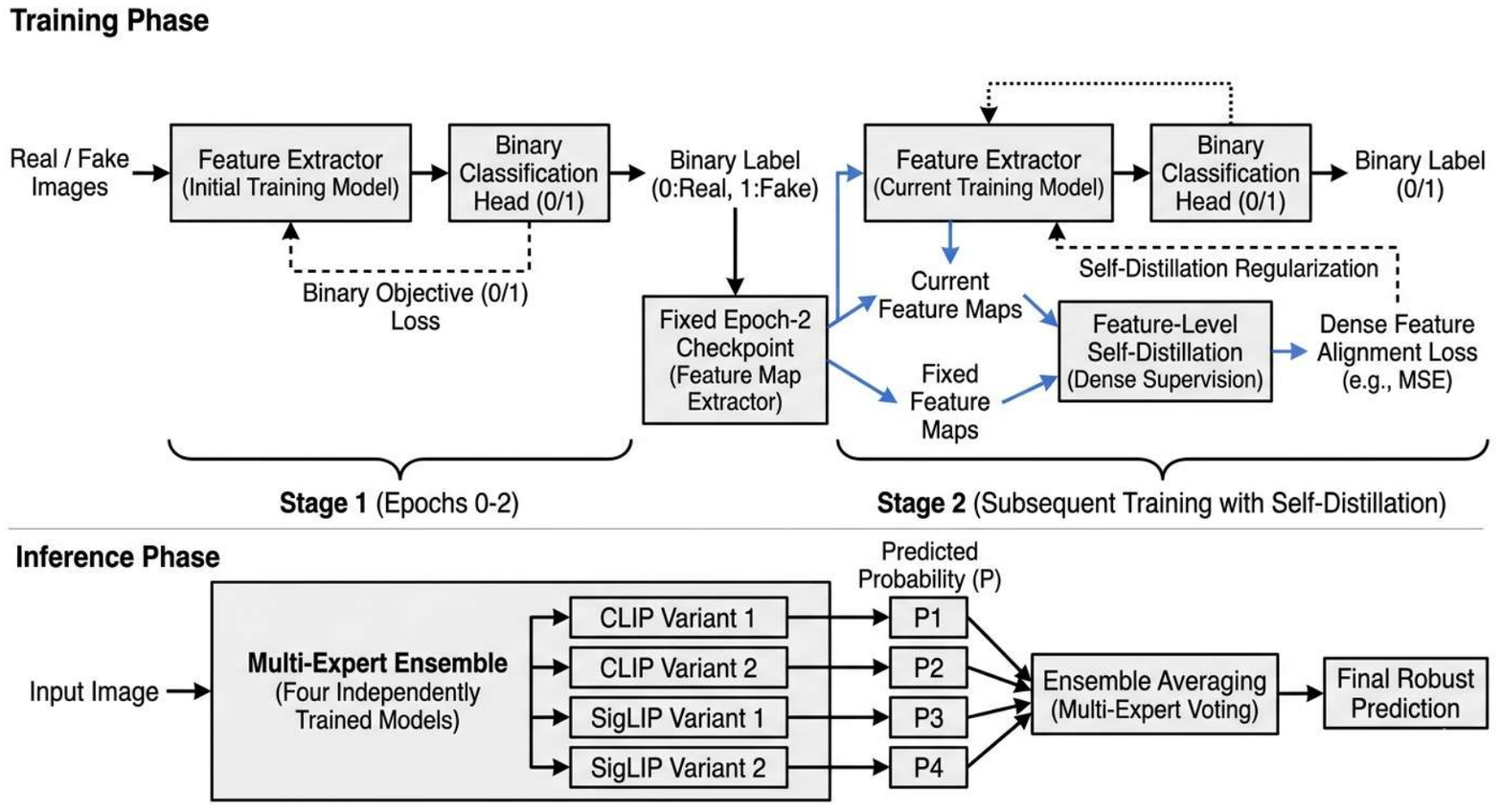}
    \caption{UESTC method scheme.}
    \label{UESTC_scheme}
\end{figure}

\paragraph{Overview.}
Our final detector is a multi-expert ensemble composed of four Vision Transformer backbones: two CLIP ViT-L/14 models~\cite{radford2021clip} and two SigLIP So400M Patch14-384 models~\cite{zhai2023siglip}.  
For each backbone, we attach a lightweight classification head that maps high-dimensional feature representations to binary predictions.  
Because the backbones have different native input configurations, we use backbone-specific preprocessing and standardize inputs to \(224\times224\) for CLIP and \(384\times384\) for SigLIP.  
Despite using four experts, inference remains efficient, with a peak GPU memory footprint of approximately 10\,GB.

\paragraph{Backbone selection.}
To identify suitable feature extractors, we conduct a large-scale empirical study over diverse backbones, including CLIP~\cite{radford2021clip}, SigLIP~\cite{zhai2023siglip}, MoCo~\cite{he2020moco}, Swin Transformer~\cite{liu2021swin}, ConvNeXt~\cite{liu2022convnext}, DINO~\cite{caron2021dino}, CoCa~\cite{yu2022coca}, and BEiT~\cite{bao2021beit}.  
We compare representation quality, computational efficiency, and robustness to synthetic artifacts.  
Empirically, CLIP and SigLIP provide the best trade-off between discriminative capability and cross-domain generalization, and therefore form the basis of the final system.

\paragraph{Data strategy.}
We improve robustness through two complementary components: training-set expansion and comprehensive degradation modeling.  
First, we expand the official training set with diverse synthesis pipelines, including state-of-the-art diffusion generators~\cite{ho2020ddpm,rombach2022high}, reconstruction pipelines, domain-transfer methods, and targeted adversarial attacks.  
This expansion introduces both realistic AI-generated samples and challenging hard negatives, exposing the detector to broader failure modes and emerging forgery artifacts.  
Second, we extend the official augmentation scripts with a wider array of degradation operators, covering complex blurs, noise patterns, compression artifacts, and advanced optical/digital distortions.  
These realistic perturbations encourage degradation-invariant representations and significantly improve robustness under distribution shifts caused by imaging conditions, transmission errors, and post-processing pipelines.

\paragraph{Training paradigm.}
We treat deepfake detection as more than a vanilla binary classification task and adopt a two-stage training strategy.  
In Stage~1, each model is trained for two epochs with a standard binary objective (real vs.\ generated).  
In Stage~2, we introduce feature-level self-distillation with dense supervision: intermediate feature maps from the epoch-2 checkpoint are used as dense targets, and the model is explicitly optimized to align current representations with these distilled features.  
This dense alignment regularizes feature geometry, reduces overfitting to the competition distribution, and improves generalization to unseen generation methods and real-world domains.

\paragraph{Inference.}
At inference time, we average the predicted probabilities from four independently trained experts (two CLIP variants and two SigLIP variants) to produce the final robust prediction.

\subsection{PSU. PRISM: Paradigm-diverse Representation Integration for Synthesis-artifact Manifold Detection}
\label{subsec:prism}

\begin{figure}[t]
    \centering
    \includegraphics[width=\columnwidth]{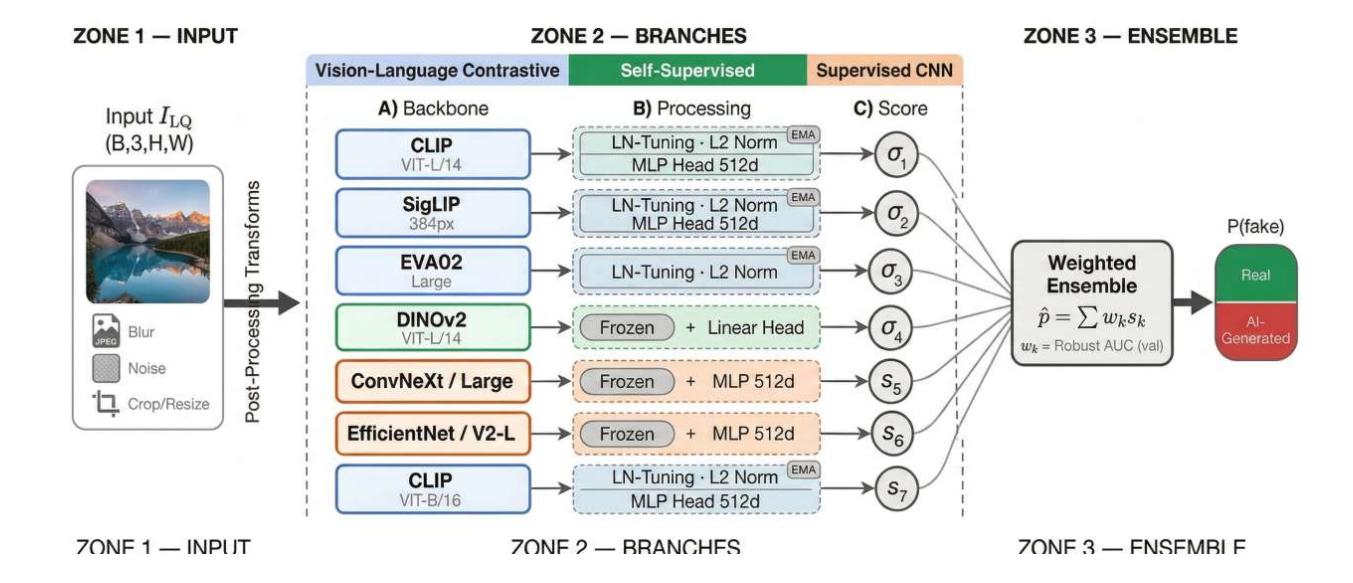}
    \caption{PSU method scheme.}
    \label{PSU_scheme}
\end{figure}

\subsubsection{Method Description}
\label{subsubsec:prism_method}

\textbf{PRISM} is a heterogeneous ensemble detector for AI-generated images, designed for robustness under JPEG compression, blur, noise, rescaling, and cropping. Our core hypothesis is that no single pre-training objective encodes the full forensic artifact manifold~\cite{wang2020cnn,corvi2023diffusiondetect}: contrastive objectives capture semantic inconsistencies; self-supervised patch objectives preserve texture discontinuities; supervised CNNs encode low-frequency spectral anomalies. Robust detection therefore requires \emph{explicit paradigm diversity}.

\paragraph{Prior work.}
Wang \textit{et al.}~\cite{wang2020cnn} showed CNN detectors generalize across GAN generators with augmentation, yet collapse under JPEG re-compression~\cite{gragnaniello2021critical}. Ojha \textit{et al.}~\cite{ojha2023towards} froze CLIP and trained a linear probe, achieving strong cross-generator generalization. PRISM extends this by (i) covering three paradigms beyond CLIP, (ii) paradigm-specific fine-tuning, and (iii) weighting by \emph{robust} AUC rather than clean accuracy.

\paragraph{Encoder pool.}
We instantiate $K{=}7$ encoders $\{\phi_k\}_{k=1}^K$ across three paradigms (Table~\ref{tab:prism_encoders}). For vision-language (VL) encoders (CLIP~\cite{radford2021clip}, SigLIP~\cite{zhai2023siglip}, EVA02~\cite{fang2024eva02}), features are $\ell_2$-normalized onto the unit hypersphere~\cite{wang2020alignmentuniformity}:
\[
\hat{z}_k = \frac{z_k}{\|z_k\|_2} \in \mathcal{S}^{d_k-1},
\]
preserving the contrastive geometry. DINOv2~\cite{oquab2023dinov2}, ConvNeXt~\cite{liu2022convnext}, and EfficientNet-V2~\cite{tan2021efficientnetv2} remain frozen; their features are passed directly to the classification head.

\paragraph{Paradigm-aware tuning.}
VL encoders undergo \emph{LayerNorm tuning}~\cite{wortsman2022robustfinetuning}: only LayerNorm scale/shift (about $0.03\%$ of weights) are updated, preventing catastrophic forgetting while adapting internal normalization to the forensic domain. All models use EMA ($\alpha{=}0.9995$) for checkpoint stabilization and calibrated posteriors. VL heads are two-layer MLPs ($h{=}512$); DINOv2 uses a linear probe. Training uses label-smoothed BCE ($\varepsilon{=}0.05$), AdamW ($\lambda{=}0.01$), cosine LR schedule, and fp16.

\begin{table}[t]
\centering
\small
\caption{PRISM encoder pool. $d$: feature dimension. LN: LayerNorm-only tuning. F: frozen encoder.}
\label{tab:prism_encoders}
\setlength{\tabcolsep}{3pt}   
\renewcommand{\arraystretch}{1.0}
\begin{tabular}{l l c c c}
\toprule
\textbf{Encoder} & \textbf{Paradigm} & \textbf{$d$} & \textbf{Tune} & \textbf{Head} \\
\midrule
CLIP ViT-L/14         & VL-Contrastive    & 1024 & LN & MLP-512 \\
SigLIP ViT-L/16       & VL-Contrastive    & 1024 & LN & MLP-512 \\
EVA02-Large           & VL-Contrastive    & 1024 & LN & MLP-512 \\
DINOv2 ViT-L/14       & Self-Supervised   & 1024 & F  & Linear \\
ConvNeXt-Large        & Supervised CNN    & 1536 & F  & MLP-512 \\
EfficientNet-V2-L     & Supervised CNN    & 1280 & F  & MLP-512 \\
CLIP ViT-B/16         & VL-Contrastive    & 512  & LN & MLP-256 \\
\bottomrule
\end{tabular}
\end{table}

\paragraph{Robust ensemble.}
Each model is scored on held-out validation data under all degradation types, yielding per-model robust AUC $A_k^{\mathrm{rob}}$. Normalized weights
\[
w_k = \frac{A_k^{\mathrm{rob}}}{\sum_j A_j^{\mathrm{rob}}}
\]
drive the ensemble. The final prediction with horizontal-flip TTA is
\begin{equation}
\hat{p} = \frac{1}{2}\left[\sum_k w_k\,p_k(x) + \sum_k w_k\,p_k(\mathcal{F}(x))\right].
\label{eq:prism_ensemble}
\end{equation}
Early stopping maximizes
\[
\mathcal{M} = 0.7A^{\text{rob}} + 0.3A^{\text{clean}}.
\]

\subsection{Shallow Real}

\subsubsection{Method Description}
\label{subsubsec:ntire2026_description}

\begin{figure}[t]
    \centering
    \includegraphics[width=\linewidth]{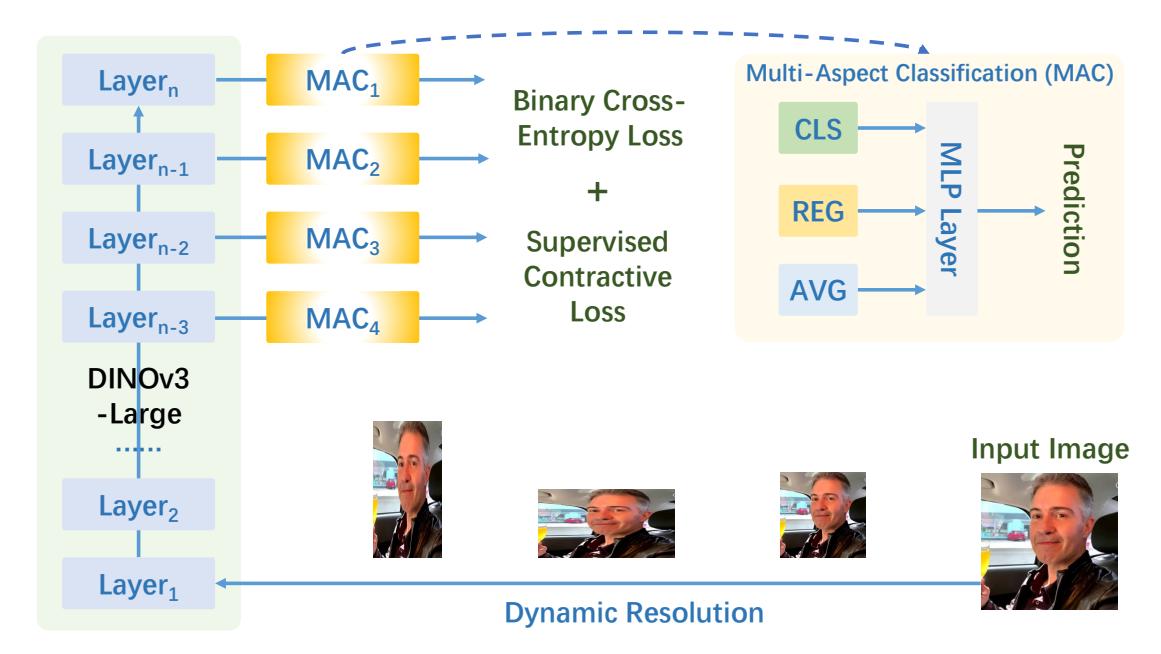}
    \caption{Shallow Real method scheme. Overall pipeline of the proposed model.}
    \label{fig:ntire2026_pipeline}
\end{figure}

As shown in Fig.~\ref{fig:ntire2026_pipeline}, we formulate robust AI-generated image detection as a binary classification task. Our model uses a DINOv3-Large backbone (following the DINO line of self-supervised visual encoders~\cite{oquab2023dinov2}), fine-tuned with Low-Rank Adaptation (LoRA)~\cite{hu2022lora} using rank $r=32$ and scaling factor $\alpha=64$.

Final prediction is produced by a Multi-Aspect Classification (MAC) head operating on DINOv3 features. Specifically, the head aggregates:
(i) one \texttt{[CLS]} token,
(ii) four \texttt{[REG]} tokens, and
(iii) one \texttt{[AVG]} token computed as the average of the remaining patch tokens.
These six 1024-dimensional vectors are concatenated into a 6144-dimensional representation and passed to a two-layer MLP with an intermediate ReLU activation to output the binary score.

During training, we use three key techniques:
\begin{enumerate}
    \item \textbf{Dynamic resolution.}
    Input images are resized to random resolutions, where both height and width are sampled from $[384, 1152]$ (multiples of patch size 16), with a randomly selected interpolation method.

    \item \textbf{Deep supervision.}
    Auxiliary classification heads are attached to the last four layers of the DINOv3 backbone. Training loss is computed for each auxiliary output, while only the final-layer output is used at inference time. Each auxiliary head uses dropout with rate 0.2 in its fully connected layer.

    \item \textbf{Metric learning objective.}
    A supervised contrastive loss~\cite{khosla2020supcon} is added as an auxiliary objective alongside the primary binary cross-entropy loss, improving feature discrimination.
\end{enumerate}

The model is trained for 10 epochs using AdamW~\cite{loshchilov2017decoupled} with a linear learning-rate decay from $10^{-4}$ to $0$. Training uses the union of the official training set and the CommunityForensics dataset, with batch size 32 and the official data-augmentation pipeline.

\begin{table*}[h!]
\nohyphens{%
\centering
\small
\setlength{\tabcolsep}{4pt}
\renewcommand{\arraystretch}{1.2}
\begin{threeparttable}
\begin{tabularx}{\textwidth}{@{}l c c c X X@{}}
\toprule
\textbf{Split} & \makecell{\textbf{\# of}\\\textbf{Images}} & \makecell{\textbf{Real/Fake}\\ \textbf{Ratio}} & \makecell{\textbf{Labels}\\ \textbf{Provided}} & \makecell{\centering \textbf{Generator Models}} & \makecell{\centering \textbf{Transformations}} \\
\midrule

\textbf{Train} &
$\sim$277K &
$\sim$1:1.77 &
Yes &
\makecell{\centering \textbf{20 Models:}}
\vspace{-0.8em}
\begin{multicols}{2}
\begin{itemize}[leftmargin=*,nosep]
\item YOSO PixArt-512
\item PixArt-$\alpha$
\item PixArt-$\Sigma$
\item Kandinsky 2
\item Kandinsky 3
\item Kolors
\item OmniGen
\item OmniGen 2
\item Stable Diffusion 1.4
\item Stable Diffusion 1.5
\item Stable Diffusion 2.1

\item Stable Diffusion XL 1.0
\item SDXL Lightning
\item SDXL Turbo
\item Janus Pro 7B
\item Infinity 2B
\item Infinity 8B
\item Ovis Image
\item DeepFloyd IF
\item FLUX.1 Kontext Dev
\end{itemize}
\end{multicols}
&
\makecell{\centering \textbf{12 Transformations:} }
\newline
\makecell{\centering \textit{(Provided as distortion pipeline)}}

\begin{multicols}{2}
\begin{itemize}[leftmargin=*,nosep]
\item Gaussian Blur
\item Lens Blur
\item Color Shift
\item Color Saturation
\item JPEG Compression
\item White Noise
\item Impulse Noise
\item Brightness Increase
\item Brightness Decrease
\item Color Jitter
\item Color Quantization
\item Linear Contrast Change
\end{itemize}
\end{multicols}
\\

\midrule

\textbf{Validation} &
10K &
1:1 &
\makecell{Yes\\ \textit{(After} \\ \textit{Completion)}}
&
\makecell{\centering \textbf{9 Models:}}
\vspace{-0.8em}
\begin{multicols}{2}
\begin{itemize}[leftmargin=*,nosep]
\item FLUX.1 Kontext Dev
\item SDXL Turbo
\item FLUX.1 Dev
\item Playground v2.5
\item Lumina Image 2.0
\item Qwen Image
\item Stable Diffusion 3 Medium
\item Ideogram v3 Turbo\textsuperscript{\dag}
\item ImageGen-4 Fast\textsuperscript{\dag}
\end{itemize}
\end{multicols}
&
\makecell{\centering \textbf{19 Transformations:}}
\begin{multicols}{2}
\begin{itemize}[leftmargin=*,nosep]
\item Gaussian Blur
\item Lens Blur
\item Color Shift
\item Color Saturation
\item JPEG Compression
\item White Noise
\item Impulse Noise
\item Brightness Increase
\item Brightness Decrease
\item Color Jitter
\item Color Quantization
\item Linear Contrast Change
\item Motion Blur
\item Multiplicative Noise
\item Pixelation
\item RGB Channel Shift
\item Random Crop
\item Random Aspect Crop
\item Downscale
\end{itemize}
\end{multicols}
\\

\midrule

\makecell{\textbf{Validation}\\\textbf{Hard}} &
2.5K &
1:1 &
\makecell{Yes\\ \textit{(After} \\ \textit{Completion)}} &
\makecell{\centering \textbf{7 Models:}}
\vspace{-0.8em}
\begin{multicols}{2}
\begin{itemize}[leftmargin=*,nosep]
\item Playground v2.5
\item SDXL Turbo
\item HiDream
\item FLUX.1 Schnell
\item Stable Diffusion 3.5 Large Turbo
\item Nano Banana\textsuperscript{\dag}
\item Seedream 4\textsuperscript{\dag}
\end{itemize}
\end{multicols}
&
\makecell{\centering \textbf{19 Transformations:}}
\begin{multicols}{2}
\begin{itemize}[leftmargin=*,nosep]
\item Gaussian Blur
\item Lens Blur
\item JPEG Compression
\item White Noise
\item Impulse Noise
\item Color Quantization
\item Multiplicative Noise
\item RGB Channel Shift
\item Random Crop
\item Random Aspect Crop
\item Neural Image Compression (JPEG AI)
\item Random Tone Curve
\item CLAHE
\item ISO Noise
\item Perspective Transform
\item Multiple Compressions (JPEG)
\item Multiple Compressions (JPEG + JPEG AI)
\item Watermark Attack (Adv. Embedding, CLIP/ResNet)
\item Downscale
\end{itemize}
\end{multicols}
\\

\bottomrule
\end{tabularx}
\end{threeparttable}
}
\caption{Detailed information on Train, Validation, and Validation Hard dataset splits. \textsuperscript{\dag} denotes a proprietary generation model used via API.}
\label{tab:splits_1}
\end{table*}

\begin{table*}[h!]
\nohyphens{
\centering
\small
\setlength{\tabcolsep}{4pt}
\renewcommand{\arraystretch}{1.2}
\begin{threeparttable}
\begin{tabularx}{\textwidth}{@{}l c c c X X@{}}
\toprule
\textbf{Split} & \makecell{\textbf{\# of}\\\textbf{Images}} & \makecell{\textbf{Real/Fake}\\ \textbf{Ratio}} & \makecell{\textbf{Labels}\\ \textbf{Provided}} & \textbf{Generator Models} & \textbf{Transformations} \\
\midrule

\makecell{\textbf{Test}\\ \textbf{(Public)}} &
2.5K &
1:1 &
\makecell{Yes\\ \textit{(After} \\ \textit{Completion)}} &
\makecell{\centering \textbf{10 Models:}}
\vspace{-0.8em}
\begin{multicols}{2}
\begin{itemize}[leftmargin=*,nosep]
\item HiDream
\item FLUX.1 Schnell
\item Stable Diffusion 3.5 Large
\item FLUX Krea
\item Z-Image Turbo
\item Nano Banana Pro\textsuperscript{\dag}
\item FLUX-2 Max\textsuperscript{\dag}
\item ImageGen-4 Ultra\textsuperscript{\dag}
\item Seedream 5 Lite\textsuperscript{\dag}
\item Grok Imagine Image\textsuperscript{\dag}
\end{itemize}
\end{multicols}
&
\makecell{\centering \textbf{22 Transformations:}}
\begin{multicols}{2}
\begin{itemize}[leftmargin=*,nosep]
\item Color Saturation
\item Brightness Increase
\item Lens Blur
\item JPEG Compression
\item Impulse Noise
\item RGB Channel Shift
\item Random Crop
\item Random Aspect Crop
\item Neural Image Compression (JPEG AI)
\item Random Tone Curve
\item CLAHE
\item ISO Noise
\item Perspective Transform
\item Multiple Compressions (JPEG)
\item Multiple Compressions (JPEG + JPEG AI)
\item Watermark Attack (Adv. Embedding, CLIP/ResNet)
\item JPEG 2000
\item Watermark Attack (WMForger)
\item Neural Image Compression (Cheng2020)
\item Shot Noise
\item Downscale
\item Invisible Watermark Insertion (1 of 6 algorithms)
\end{itemize}
\end{multicols}
\\

\midrule

\makecell{\textbf{Test}\\ \textbf{(Private)}} &
2.5K &
$\sim$1:1 &
No &
\makecell{\centering \textbf{10 Models:}}
\vspace{-0.8em}
\begin{multicols}{2}
\begin{itemize}[leftmargin=*,nosep]
\item HiDream
\item Stable Diffusion 3.5 Large Turbo
\item FLUX.1 Dev SRPO
\item Z-Image Turbo
\item Kandinsky 5
\item Nano Banana 2\textsuperscript{\dag}
\item GPT Image 1.5\textsuperscript{\dag}
\item ImageGen-4 Ultra\textsuperscript{\dag}
\item Seedream 5 Lite\textsuperscript{\dag}
\item Grok Imagine Image\textsuperscript{\dag}
\end{itemize}
\end{multicols}
&
\makecell{\centering \textbf{24 Transformations:}}
\begin{multicols}{2}
\begin{itemize}[leftmargin=*,nosep]
\item Color Saturation
\item Brightness Increase
\item Lens Blur
\item JPEG Compression
\item Impulse Noise
\item RGB Channel Shift
\item Random Crop
\item Random Aspect Crop
\item Neural Image Compression (JPEG AI)
\item Random Tone Curve
\item CLAHE
\item ISO Noise
\item Perspective Transform
\item Multiple Compressions (JPEG)
\item Multiple Compressions (JPEG + JPEG AI)
\item Multiple Compressions (JPEG + JPEG 2000)
\item Watermark Attack (Adv. Embedding, CLIP/ResNet)
\item JPEG 2000
\item Watermark Attack (WMForger)
\item Neural Image Compression (Cheng2020)
\item Shot Noise
\item Glass Blur
\item Downscale
\item Invisible Watermark Insertion (1 of 7 algorithms)
\end{itemize}
\end{multicols}
\\

\bottomrule
\end{tabularx}
\end{threeparttable}
}
\caption{Detailed information on Test (public) and Test (private) dataset splits. \textsuperscript{\dag} denotes a proprietary generation model used via API.}
\label{tab:splits_2}
\end{table*}

{
    \small
    \bibliographystyle{ieeenat_fullname}
    \bibliography{main}

@String(CVPR= {IEEE Conf. Comput. Vis. Pattern Recog.})

@String(ICCV= {Int. Conf. Comput. Vis.})

@String(NIPS= {Adv. Neural Inform. Process. Syst.})

@String(ICME = {Int. Conf. Multimedia and Expo})

@String(ICASSP=	{ICASSP})

@String(ICLR = {Int. Conf. Learn. Represent.})

@String(AAAI = {AAAI})

@String(CVPR  = {CVPR})

@String(ICCV  = {ICCV})

@String(NIPS  = {NeurIPS})

@String(ICME  =	{ICME})

@String(ICLR  = {ICLR})

@inproceedings{rombach2022high,
  title={High-resolution image synthesis with latent diffusion models},
  author={Rombach, Robin and Blattmann, Andreas and Lorenz, Dominik and Esser, Patrick and Ommer, Bj{\"o}rn},
  booktitle={Proceedings of the IEEE/CVF conference on computer vision and pattern recognition},
  pages={10684--10695},
  year={2022}
}

@article{goodfellow2020generative,
  title={Generative adversarial networks},
  author={Goodfellow, Ian and Pouget-Abadie, Jean and Mirza, Mehdi and Xu, Bing and Warde-Farley, David and Ozair, Sherjil and Courville, Aaron and Bengio, Yoshua},
  journal={Communications of the ACM},
  volume={63},
  number={11},
  pages={139--144},
  year={2020},
  publisher={ACM New York, NY, USA}
}

@article{li2024autoregressive,
  title={Autoregressive image generation without vector quantization},
  author={Li, Tianhong and Tian, Yonglong and Li, He and Deng, Mingyang and He, Kaiming},
  journal={Advances in Neural Information Processing Systems},
  volume={37},
  pages={56424--56445},
  year={2024}
}

@inproceedings{tan2024rethinking,
  title={Rethinking the up-sampling operations in cnn-based generative network for generalizable deepfake detection},
  author={Tan, Chuangchuang and Zhao, Yao and Wei, Shikui and Gu, Guanghua and Liu, Ping and Wei, Yunchao},
  booktitle={Proceedings of the IEEE/CVF conference on computer vision and pattern recognition},
  pages={28130--28139},
  year={2024}
}

@inproceedings{frank2020leveraging,
  title={Leveraging frequency analysis for deep fake image recognition},
  author={Frank, Joel and Eisenhofer, Thorsten and Sch{\"o}nherr, Lea and Fischer, Asja and Kolossa, Dorothea and Holz, Thorsten},
  booktitle={International conference on machine learning},
  pages={3247--3258},
  year={2020},
  organization={PMLR}
}

@inproceedings{wang2020cnn,
  title={CNN-generated images are surprisingly easy to spot... for now},
  author={Wang, Sheng-Yu and Wang, Oliver and Zhang, Richard and Owens, Andrew and Efros, Alexei A},
  booktitle={Proceedings of the IEEE/CVF conference on computer vision and pattern recognition},
  pages={8695--8704},
  year={2020}
}

@inproceedings{wang2023dire,
  title={Dire for diffusion-generated image detection},
  author={Wang, Zhendong and Bao, Jianmin and Zhou, Wengang and Wang, Weilun and Hu, Hezhen and Chen, Hong and Li, Houqiang},
  booktitle={Proceedings of the IEEE/CVF International Conference on Computer Vision},
  pages={22445--22455},
  year={2023}
}

@inproceedings{ojha2023towards,
  title={Towards universal fake image detectors that generalize across generative models},
  author={Ojha, Utkarsh and Li, Yuheng and Lee, Yong Jae},
  booktitle={Proceedings of the IEEE/CVF conference on computer vision and pattern recognition},
  pages={24480--24489},
  year={2023}
}

@inproceedings{li2025bridging,
  title={Bridging the Gap Between Ideal and Real-world Evaluation: Benchmarking AI-Generated Image Detection in Challenging Scenarios},
  author={Li, Chunxiao and Wang, Xiaoxiao and Li, Meiling and Miao, Boming and Sun, Peng and Zhang, Yunjian and Ji, Xiangyang and Zhu, Yao},
  booktitle={Proceedings of the IEEE/CVF International Conference on Computer Vision},
  pages={20379--20389},
  year={2025}
}

@article{zhu2023genimage,
  title={Genimage: A million-scale benchmark for detecting ai-generated image},
  author={Zhu, Mingjian and Chen, Hanting and Yan, Qiangyu and Huang, Xudong and Lin, Guanyu and Li, Wei and Tu, Zhijun and Hu, Hailin and Hu, Jie and Wang, Yunhe},
  journal={Advances in neural information processing systems},
  volume={36},
  pages={77771--77782},
  year={2023}
}

@inproceedings{hong2025wildfake,
  title={Wildfake: A large-scale and hierarchical dataset for ai-generated images detection},
  author={Hong, Yan and Feng, Jianming and Chen, Haoxing and Lan, Jun and Zhu, Huijia and Wang, Weiqiang and Zhang, Jianfu},
  booktitle={Proceedings of the AAAI Conference on Artificial Intelligence},
  volume={39},
  number={4},
  pages={3500--3508},
  year={2025}
}

@article{yan2024df40,
  title={Df40: Toward next-generation deepfake detection},
  author={Yan, Zhiyuan and Yao, Taiping and Chen, Shen and Zhao, Yandan and Fu, Xinghe and Zhu, Junwei and Luo, Donghao and Wang, Chengjie and Ding, Shouhong and Wu, Yunsheng and others},
  journal={Advances in Neural Information Processing Systems},
  volume={37},
  pages={29387--29434},
  year={2024}
}

@inproceedings{hu2022lora,
  title     = {LoRA: Low-Rank Adaptation of Large Language Models},
  author    = {Hu, Edward J and Shen, Yelong and Wallis, Phillip and Allen-Zhu, Zeyuan and Li, Yuanzhi and Wang, Shean and Wang, Liang and Chen, Weizhu},
  booktitle = {International Conference on Learning Representations (ICLR)},
  year      = {2022}
}

@article{sun2023evaclip,
  title   = {EVA-CLIP: Improved Training Techniques for CLIP at Scale},
  author  = {Sun, Quan and Fang, Yuxin and Wu, Ledell and Wang, Xinlong and Cao, Yue},
  journal = {arXiv preprint arXiv:2303.15389},
  year    = {2023}
}

@article{huang2025sofake,
  title   = {So-Fake: Benchmarking and Explaining Social Media Image Forgery Detection},
  author  = {Huang, Zhenglin and Li, Tianxiao and Li, Xiangtai and Wen, Haiquan and He, Yiwei and Zhang, Jiangning and Fei, Hao and Yang, Xi and Huang, Xiaowei and Peng, Bei and others},
  journal = {arXiv preprint arXiv:2505.18660},
  year    = {2025}
}

@article{loshchilov2016sgdr,
  title   = {SGDR: Stochastic Gradient Descent with Warm Restarts},
  author  = {Loshchilov, Ilya and Hutter, Frank},
  journal = {arXiv preprint arXiv:1608.03983},
  year    = {2016}
}

@article{loshchilov2017decoupled,
  title   = {Decoupled Weight Decay Regularization},
  author  = {Loshchilov, Ilya and Hutter, Frank},
  journal = {arXiv preprint arXiv:1711.05101},
  year    = {2017}
}

@article{yan2024sanity,
  title   = {A Sanity Check for AI-Generated Image Detection},
  author  = {Yan, Shilin and Li, Ouxiang and Cai, Jiayin and Hao, Yanbin and Jiang, Xiaolong and Hu, Yao and Xie, Weidi},
  journal = {arXiv preprint arXiv:2406.19435},
  year    = {2024}
}

@misc{siglip2,
  title        = {SigLIP 2: Multilingual Vision-Language Encoders with Improved Semantic Understanding, Localization, and Dense Features},
  author       = {Google},
  year         = {2025},
  howpublished = {\url{https://huggingface.co/google/siglip2-giant-opt-patch16-384}},
  note         = {Model card / release page, accessed: 2026-03-20}
}

@inproceedings{radford2021clip,
  title     = {Learning Transferable Visual Models From Natural Language Supervision},
  author    = {Radford, Alec and Kim, Jong Wook and Hallacy, Chris and Ramesh, Aditya and Goh, Gabriel and Agarwal, Sandhini and Sastry, Girish and Askell, Amanda and Mishkin, Pamela and Clark, Jack and Krueger, Gretchen and Sutskever, Ilya},
  booktitle = {International Conference on Machine Learning (ICML)},
  pages     = {8748--8763},
  year      = {2021}
}

@inproceedings{zhai2023siglip,
  title     = {Sigmoid Loss for Language Image Pre-Training},
  author    = {Zhai, Xiaohua and Mustafa, Basil and Kolesnikov, Alexander and Beyer, Lucas},
  booktitle = {Proceedings of the IEEE/CVF International Conference on Computer Vision (ICCV)},
  pages     = {11975--11986},
  year      = {2023}
}

@inproceedings{fang2024eva02,
  title     = {EVA-02: A Visual Representation for Neon Genesis},
  author    = {Fang, Yuxin and Wang, Wenhai and Xie, Binhui and Sun, Quan and Wu, Ledell and Wang, Xinggang and Cao, Yue and others},
  booktitle = {Proceedings of the IEEE/CVF Conference on Computer Vision and Pattern Recognition (CVPR)},
  year      = {2024}
}

@article{fridrich2012rich,
  title   = {Rich Models for Steganalysis of Digital Images},
  author  = {Fridrich, Jessica and Kodovsk{\'y}, Jan},
  journal = {IEEE Transactions on Information Forensics and Security},
  volume  = {7},
  number  = {3},
  pages   = {868--882},
  year    = {2012}
}

@inproceedings{arniqa,
  title={Arniqa: Learning distortion manifold for image quality assessment},
  author={Agnolucci, Lorenzo and Galteri, Leonardo and Bertini, Marco and Del Bimbo, Alberto},
  booktitle={Proceedings of the IEEE/CVF Winter Conference on Applications of Computer Vision},
  pages={189--198},
  year={2024}
}

@misc{nanobanana2,
  title = {Nano Banana 2 image generation model from Google},
  howpublished = {\url{https://blog.google/innovation-and-ai/technology/ai/nano-banana-2/}},
  note = {Accessed: 2026-03-22},
  key = {none}
}

@misc{seedream5,
  title = {SeeDream 5 Lite image generation model from ByteDance},
  howpublished = {\url{https://seed.bytedance.com/en/seedream5_0_lite}},
  note = {Accessed: 2026-03-22},
  key = {none}
}

@inproceedings{changpinyo2021conceptual,
  title={Conceptual 12m: Pushing web-scale image-text pre-training to recognize long-tail visual concepts},
  author={Changpinyo, Soravit and Sharma, Piyush and Ding, Nan and Soricut, Radu},
  booktitle={Proceedings of the IEEE/CVF conference on computer vision and pattern recognition},
  pages={3558--3568},
  year={2021}
}

@article{gadre2023datacomp,
  title={Datacomp: In search of the next generation of multimodal datasets},
  author={Gadre, Samir Yitzhak and Ilharco, Gabriel and Fang, Alex and Hayase, Jonathan and Smyrnis, Georgios and Nguyen, Thao and Marten, Ryan and Wortsman, Mitchell and Ghosh, Dhruba and Zhang, Jieyu and others},
  journal={Advances in Neural Information Processing Systems},
  volume={36},
  pages={27092--27112},
  year={2023}
}

@article{desai2021redcaps,
  title={Redcaps: Web-curated image-text data created by the people, for the people},
  author={Desai, Karan and Kaul, Gaurav and Aysola, Zubin and Johnson, Justin},
  journal={arXiv preprint arXiv:2111.11431},
  year={2021}
}

@article{labs2025flux,
  title={FLUX. 1 Kontext: Flow Matching for In-Context Image Generation and Editing in Latent Space},
  author={Labs, Black Forest and Batifol, Stephen and Blattmann, Andreas and Boesel, Frederic and Consul, Saksham and Diagne, Cyril and Dockhorn, Tim and English, Jack and English, Zion and Esser, Patrick and others},
  journal={arXiv preprint arXiv:2506.15742},
  year={2025}
}

@misc{deepfloyd,
  title = {IF image generation model from DeepFloyd Lab},
  howpublished = {\url{https://github.com/deep-floyd/IF}},
  note = {Accessed: 2026-03-22},
  key = {none}
}

@article{wang2025ovis_image,
  title={Ovis-Image Technical Report}, 
  author={Wang, Guo-Hua and Cao, Liangfu and Cui, Tianyu and Fu, Minghao and Chen, Xiaohao and Zhan, Pengxin and Zhao, Jianshan and Li, Lan and Fu, Bowen and Liu, Jiaqi and Chen, Qing-Guo},
  journal={arXiv preprint arXiv:2511.22982},
  year={2025}
}

@article{wu2025qwen,
  title={Qwen-image technical report},
  author={Wu, Chenfei and Li, Jiahao and Zhou, Jingren and Lin, Junyang and Gao, Kaiyuan and Yan, Kun and Yin, Sheng-ming and Bai, Shuai and Xu, Xiao and Chen, Yilei and others},
  journal={arXiv preprint arXiv:2508.02324},
  year={2025}
}

@article{hidreami1technicalreport,
  title={HiDream-I1: A High-Efficient Image Generative Foundation Model with Sparse Diffusion Transformer},
  author={Cai, Qi and Chen, Jingwen and Chen, Yang and Li, Yehao and Long, Fuchen and Pan, Yingwei and Qiu, Zhaofan and Zhang, Yiheng and Gao, Fengbin and Xu, Peihan and others},
  journal={arXiv preprint arXiv:2505.22705},
  year={2025}
}

@misc{nanobanana,
  title = {Nano Banana image generation model from Google},
  howpublished = {\url{https://blog.google/products-and-platforms/products/gemini/updated-image-editing-model/}},
  note = {Accessed: 2026-03-22},
  key = {none}
}

@misc{grok,
  title = {Grok Imagine image generation model from XAI},
  howpublished = {\url{https://grok.com/imagine}},
  note = {Accessed: 2026-03-22},
  key = {none}
}

@inproceedings{he2020moco,
  title     = {Momentum Contrast for Unsupervised Visual Representation Learning},
  author    = {He, Kaiming and Fan, Haoqi and Wu, Yuxin and Xie, Saining and Girshick, Ross},
  booktitle = CVPR,
  pages     = {9729--9738},
  year      = {2020}
}

@inproceedings{liu2021swin,
  title     = {Swin Transformer: Hierarchical Vision Transformer Using Shifted Windows},
  author    = {Liu, Ze and Lin, Yutong and Cao, Yue and Hu, Han and Wei, Yixuan and Zhang, Zheng and Lin, Stephen and Guo, Baining},
  booktitle = ICCV,
  pages     = {10012--10022},
  year      = {2021}
}

@inproceedings{liu2022convnext,
  title     = {A ConvNet for the 2020s},
  author    = {Liu, Zhuang and Mao, Hanzi and Wu, Chao-Yuan and Feichtenhofer, Christoph and Darrell, Trevor and Xie, Saining},
  booktitle = CVPR,
  pages     = {11976--11986},
  year      = {2022}
}

@inproceedings{caron2021dino,
  title     = {Emerging Properties in Self-Supervised Vision Transformers},
  author    = {Caron, Mathilde and Touvron, Hugo and Misra, Ishan and J{\'e}gou, Herv{\'e} and Mairal, Julien and Bojanowski, Piotr and Joulin, Armand},
  booktitle = ICCV,
  pages     = {9650--9660},
  year      = {2021}
}

@article{yu2022coca,
  title   = {CoCa: Contrastive Captioners are Image-Text Foundation Models},
  author  = {Yu, Jiahui and Wang, Wenhu and Vasudevan, Vijay and Yeung, Leung and Seyedhosseini, Mojtaba and Wu, Yonghui},
  journal = {Transactions on Machine Learning Research},
  year    = {2022}
}

@article{bao2021beit,
  title   = {BEiT: BERT Pre-Training of Image Transformers},
  author  = {Bao, Hangbo and Dong, Li and Wei, Furu},
  journal = {arXiv preprint arXiv:2106.08254},
  year    = {2021}
}

@inproceedings{ho2020ddpm,
  title     = {Denoising Diffusion Probabilistic Models},
  author    = {Ho, Jonathan and Jain, Ajay and Abbeel, Pieter},
  booktitle = NIPS,
  pages     = {6840--6851},
  year      = {2020}
}

@inproceedings{corvi2023diffusiondetect,
  title     = {On the Detection of Synthetic Images Generated by Diffusion Models},
  author    = {Corvi, Riccardo and Cozzolino, Davide and Zingarini, Guido and Poggi, Giovanni and Nagano, Kazuhiro and Verdoliva, Luisa},
  booktitle = ICASSP,
  pages     = {1--5},
  year      = {2023}
}

@inproceedings{gragnaniello2021critical,
  title     = {Are GAN Generated Images Easy to Detect? A Critical Analysis of the State-of-the-Art},
  author    = {Gragnaniello, Diego and Cozzolino, Davide and Marra, Francesco and Poggi, Giovanni and Verdoliva, Luisa},
  booktitle = ICME,
  pages     = {1--6},
  year      = {2021}
}

@article{wang2020alignmentuniformity,
  title   = {Understanding Contrastive Representation Learning through Alignment and Uniformity on the Hypersphere},
  author  = {Wang, Tongzhou and Isola, Phillip},
  journal = {arXiv preprint arXiv:2005.10242},
  year    = {2020}
}

@article{oquab2023dinov2,
  title   = {DINOv2: Learning Robust Visual Features without Supervision},
  author  = {Oquab, Maxime and Darcet, Timoth{\'e}e and Moutakanni, Th{\'e}o and Vo, Huy V. and Szafraniec, Wiktor and Khalidov, Vasil and Fernandez, Pierre and Haziza, Daniel and Massa, Francisco and El-Nouby, Alaa and Assran, Mahmoud and Ballas, Nicolas and Galuba, Wojciech and Howes, Russell and Huang, Po-Yao and Li, Shang-Wen and Misra, Ishan and Rabbat, Michael and Sharma, Vasu and Synnaeve, Gabriel and Xu, Hu and J{\'e}gou, Herv{\'e} and Mairal, Julien and Labatut, Patrick and Joulin, Armand and Bojanowski, Piotr},
  journal = {arXiv preprint arXiv:2304.07193},
  year    = {2023}
}

@inproceedings{tan2021efficientnetv2,
  title     = {EfficientNetV2: Smaller Models and Faster Training},
  author    = {Tan, Mingxing and Le, Quoc V.},
  booktitle = {ICML},
  pages     = {10096--10106},
  year      = {2021}
}

@article{wortsman2022robustfinetuning,
  title   = {Robust Fine-Tuning of Zero-Shot Models},
  author  = {Wortsman, Mitchell and Ilharco, Gabriel and Li, Mike and Kim, Jong Wook and Hajishirzi, Hannaneh and Farhadi, Ali and Namkoong, Hongseok and Schmidt, Ludwig},
  journal = {arXiv preprint arXiv:2109.01903},
  year    = {2022}
}

@inproceedings{khosla2020supcon,
  title     = {Supervised Contrastive Learning},
  author    = {Khosla, Prannay and Teterwak, Piotr and Wang, Chen and Sarna, Aaron and Tian, Yonglong and Isola, Phillip and Maschinot, Aaron and Liu, Ce and Krishnan, Dilip},
  booktitle = NIPS,
  pages     = {18661--18673},
  year      = {2020}
}

@inproceedings{yakushev2025wibe,
	title={WIBE: Watermarks for generated Images-Benchmarking \& Evaluation},
	author={Yakushev, A and Akimenkov, A and Abud, K and Obydenkov, D and Serzhenko, I and Aistov, K and Kovalev, E and Fomin, S and Antsiferova, A and Lukianov, K and others},
	booktitle={Proceedings of the 40th IEEE/ACM International Conference on Automated Software Engineering},
	year={2025}
}

@inproceedings{cheng2020learned,
  title={Learned image compression with discretized gaussian mixture likelihoods and attention modules},
  author={Cheng, Zhengxue and Sun, Heming and Takeuchi, Masaru and Katto, Jiro},
  booktitle={Proceedings of CVPR},
  pages={7939--7948},
  year={2020}
}

@ARTICLE{jpeg_ai_standard,
  author={Ascenso, João and Alshina, Elena and Ebrahimi, Touradj},
  journal={IEEE MultiMedia}, 
  title={The JPEG AI Standard: Providing Efficient Human and Machine Visual Data Consumption}, 
  year={2023},
  volume={30},
  number={1},
  pages={100-111},
  doi={10.1109/MMUL.2023.3245919}}

@inproceedings{soucek2025wmforger,
  title={Transferable Black-Box One-Shot Forging of Watermarks via Image Preference Models},
  author={Sou\v{c}ek, Tom\'{a}\v{s} and Rebuffi, Sylvestre-Alvise and Fernandez, Pierre and Jovanović, Nikola and Elsahar, Hady and Lacatusu, Valeriu and Tran, Tuan and Mourachko, Alexandre},
  booktitle={Advances in Neural Information Processing Systems},
  year={2025}
}

@inproceedings{ntire26deepfake, 
title={{    Robust Deepfake Detection, NTIRE 2026 Challenge: Report    }}, 
author={    Hopf, Benedikt and  Timofte, Radu and others    },   
booktitle={Proceedings of the IEEE/CVF Conference on Computer Vision and Pattern Recognition (CVPR) Workshops},  
year = {2026} 
}

@inproceedings{ntire26hrdepth, 
title={{    NTIRE 2026 Challenge on High-Resolution Depth of non-Lambertian Surfaces    }}, 
author={    Zama Ramirez, Pierluigi and  Tosi, Fabio and  Di Stefano, Luigi and  Timofte, Radu and  Costanzino, Alex and  Poggi, Matteo and  Salti, Samuele and  Mattoccia, Stefano and others    },   
booktitle={Proceedings of the IEEE/CVF Conference on Computer Vision and Pattern Recognition (CVPR) Workshops},  
year = {2026} 
}

@inproceedings{ntire26raim_fusion, 
title={{    NTIRE 2026 The 3rd Restore Any Image Model (RAIM) Challenge: Multi-Exposure Image Fusion in Dynamic Scenes (Track2)    }}, 
author={    Qu, Lishen and  Liu, Yao and  Liang, Jie and  Zeng, Hui and  Dai, Wen and  Guan, Ya-nan and  Qin, Guanyi and  Zhou, Shihao and  Yang, Jufeng and  Zhang, Lei and  Timofte, Radu and others    },   
booktitle={Proceedings of the IEEE/CVF Conference on Computer Vision and Pattern Recognition (CVPR) Workshops},  
year = {2026} 
}

@inproceedings{ntire26raim_portrait, 
title={{    NTIRE 2026 The 3rd Restore Any Image Model (RAIM) Challenge: AI Flash Portrait (Track 3)    }}, 
author={    Guan, Ya-nan and  Zhang, Shaonan and  Guo, Hang and  Wang, Yawen and  Fan, Xinying and  Liang, Jie and  Zeng, Hui and  Qin, Guanyi and  Qu, Lishen and  Dai, Tao and  Xia, Shu-Tao and  Zhang, Lei and  Timofte, Radu and others    },   
booktitle={Proceedings of the IEEE/CVF Conference on Computer Vision and Pattern Recognition (CVPR) Workshops},  
year = {2026} 
}

@inproceedings{ntire26raim_piqa, 
title={{    NTIRE 2026 The 3rd Restore Any Image Model (RAIM) Challenge: Professional Image Quality Assessment (Track 1)    }}, 
author={    Qin, Guanyi and  Liang, Jie and  Zhang, Bingbing and  Qu, Lishen and  Guan, Ya-nan and  Zeng, Hui and  Zhang, Lei and  Timofte, Radu and others    },   
booktitle={Proceedings of the IEEE/CVF Conference on Computer Vision and Pattern Recognition (CVPR) Workshops},  
year = {2026} 
}

@inproceedings{ntire26lightsr, 
title={{    NTIRE 2026 Challenge on Light Field Image Super-Resolution: Methods and Results    }}, 
author={    Wang, Yingqian and  Liang, Zhengyu and  Zhang, Fengyuan and  Zhao, Wending and  Wang, Longguang and  Li, Juncheng and  Yang, Jungang and  Timofte, Radu and  Guo, Yulan and others    },   
booktitle={Proceedings of the IEEE/CVF Conference on Computer Vision and Pattern Recognition (CVPR) Workshops},  
year = {2026} 
}

@inproceedings{ntire263dsr, 
title={{    NTIRE 2026 Challenge on 3D Content Super-Resolution: Methods and Results    }}, 
author={    Wang, Longguang and  Guo, Yulan and  Wang, Yingqian and  Li, Juncheng and  Peng, Sida and  Zhang, Ye and  Timofte, Radu and  Chen, Minglin and  Wang, Yi and  Hu, Qibin and  Lei, Wenjie and others    },   
booktitle={Proceedings of the IEEE/CVF Conference on Computer Vision and Pattern Recognition (CVPR) Workshops},  
year = {2026} 
}

@inproceedings{ntire26videores, 
title={{    NTIRE 2026 Challenge on Bitstream-Corrupted Video Restoration: Methods and Results    }}, 
author={    Zou, Wenbin and  Liu, Tianyi and  Wu, Kejun and  Zhuang, Huiping and  Wu, Zongwei and  Zhou, Zhuyun and  Timofte, Radu and  others     },   booktitle={Proceedings of the IEEE/CVF Conference on Computer Vision and Pattern Recognition (CVPR) Workshops},  
year = {2026} 
}

@inproceedings{ntire26XAIGCqa, 
title={{    NTIRE 2026 X-AIGC Quality Assessment Challenge: Methods and Results    }}, 
author={    Liu, Xiaohong and  Min, Xiongkuo and  Zhai, Guangtao and  Hu, Qiang and  Cao, Jiezhang and  Zhou, Yu and  Sun, Wei and  Wen, Farong and  Xu, Zitong and  Zhou, Yingjie and  Duan, Huiyu and  Liu, Lu and  Wang, Jiarui and  Luo, Siqi and  Li, Chunyi and  Xu, Li and  Zhang, Zicheng and  Shi, Yue and  Wang, Yubo and  Zhang, Minghong and  Guo, Chunchao and  Hu, Zhichao and  Chen, Mingtao and  Wu, Xiele and  Ma, Xin and  Lv, Zhaohe and  Xue, Yuanhao and  Wang, Jiaqi and  Sha, Xinxing and  Timofte, Radu and  others    },   
booktitle={Proceedings of the IEEE/CVF Conference on Computer Vision and Pattern Recognition (CVPR) Workshops},  
year = {2026} 
}

@inproceedings{ntire26shadow, 
title={{    Advances in Single-Image Shadow Removal: Results from the NTIRE 2026 Challenge    }}, 
author={    Vasluianu, Florin-Alexandru and  Seizinger, Tim and  Zhou, Zhuyun and  Wu, Zongwei and  Timofte, Radu and  others     },   
booktitle={Proceedings of the IEEE/CVF Conference on Computer Vision and Pattern Recognition (CVPR) Workshops},  
year = {2026} 
}

@inproceedings{ntire26lightnorm, 
title={{    Learning-Based Ambient Lighting Normalization: NTIRE 2026 Challenge Results and Findings    }}, 
author={    Vasluianu, Florin-Alexandru and  Seizinger, Tim and  Chen, Jeffrey and  Zhou, Zhuyun and  Wu, Zongwei and  Timofte, Radu and  others    },   booktitle={Proceedings of the IEEE/CVF Conference on Computer Vision and Pattern Recognition (CVPR) Workshops},  
year = {2026} 
}

@inproceedings{ntire26bokeh, 
title={{    The First Controllable Bokeh Rendering Challenge at NTIRE 2026    }}, 
author={    Seizinger, Tim and  Vasluianu, Florin-Alexandru and  Conde, Marcos V. and  Chen, Jeffrey and  Zhou, Zhuyun and  Wu, Zongwei and  Timofte, Radu and  others    },   
booktitle={Proceedings of the IEEE/CVF Conference on Computer Vision and Pattern Recognition (CVPR) Workshops},  
year = {2026} 
}

@inproceedings{ntire26ripdetseg, 
title={{    NTIRE 2026 Rip Current Detection and Segmentation (RipDetSeg) Challenge Report    }}, 
author={    Dumitriu, Andrei and  Ralhan, Aakash and  Miron, Florin and  Tatui, Florin and  Ionescu, Radu Tudor and  Timofte, Radu and  others     },   booktitle={Proceedings of the IEEE/CVF Conference on Computer Vision and Pattern Recognition (CVPR) Workshops},  
year = {2026} 
}

@inproceedings{ntire26llie, 
title={{    Low Light Image Enhancement Challenge at NTIRE 2026    }}, 
author={    Ciubotariu, George and  S M A,  Sharif and  Rehman, Abdur and  Ali Dharejo, Fayaz and  Naqvi, Rizwan Ali and  Conde, Marcos and  Timofte, Radu and others    },   
booktitle={Proceedings of the IEEE/CVF Conference on Computer Vision and Pattern Recognition (CVPR) Workshops},  
year = {2026} 
}

@inproceedings{ntire26highfps, 
title={{    High FPS Video Frame Interpolation Challenge at NTIRE 2026    }}, 
author={    Ciubotariu, George and  Zhou, Zhuyun and  Jin, Yeying and  Wu, Zongwei and  Timofte, Radu and  others    },   
booktitle={Proceedings of the IEEE/CVF Conference on Computer Vision and Pattern Recognition (CVPR) Workshops},  
year = {2026} 
}

@inproceedings{ntire26nthaze, 
title={{    NT-HAZE: A Benchmark Dataset for Realistic Night-time Image Dehazing    }}, 
author={    Ancuti, Radu and  Ancuti, Codruta and  Timofte, Radu and  Ancuti, Cosmin    },   
booktitle={Proceedings of the IEEE/CVF Conference on Computer Vision and Pattern Recognition (CVPR) Workshops},  
year = {2026} 
}

@inproceedings{ntire26nthaze_rep, 
title={{    NTIRE 2026 Nighttime Image Dehazing Challenge Report    }}, 
author={    Ancuti, Radu and  Brateanu, Alexandru and  Vasluianu, Florin and  Balmez, Raul and  Orhei, Ciprian and  Ancuti, Codruta and  Timofte, Radu and  Ancuti, Cosmin and others    },   
booktitle={Proceedings of the IEEE/CVF Conference on Computer Vision and Pattern Recognition (CVPR) Workshops},  
year = {2026} 
}

@inproceedings{ntire26isp, 
title={{    NTIRE 2026 Challenge on Learned Smartphone ISP with Unpaired Data: Methods and Results    }}, 
author={    Perevozchikov, Georgy and  Vladimirov, Daniil and  Timofte, Radu and  others    },   
booktitle={Proceedings of the IEEE/CVF Conference on Computer Vision and Pattern Recognition (CVPR) Workshops},  
year = {2026} 
}

@inproceedings{ntire26ugcvideo, 
title={{    NTIRE 2026 Challenge on Short-form UGC Video Restoration in the Wild with Generative Models: Datasets, Methods and Results    }}, author={    Li, Xin and  Gong, Jiachao and  Wang, Xijun and  Xiong, Shiyao and  Li, Bingchen and  Yao, Suhang  and  Zhou, Chao and  Chen, Zhibo and  Timofte, Radu and others    },   
booktitle={Proceedings of the IEEE/CVF Conference on Computer Vision and Pattern Recognition (CVPR) Workshops},  
year = {2026} 
}

@inproceedings{ntire26dual_focus, 
title={{    NTIRE 2026 The Second Challenge on Day and Night Raindrop Removal for Dual-Focused Images: Methods and Results    }}, 
author={    Li, Xin and  Jin, Yeying and  Yao, Suhang and  Lin, Beibei and  Fan, Zhaoxin and   Yan, Wending and  Jin, Xin and  Wu, Zongwei  and  Li, Bingchen  and  Shi, Peishu and  Yang, Yufei and  Li, Yu and  Chen, Zhibo  and  Wen, Bihan and  Tan, Robby and  Timofte, Radu and others    },   
booktitle={Proceedings of the IEEE/CVF Conference on Computer Vision and Pattern Recognition (CVPR) Workshops},  
year = {2026} 
}

@inproceedings{ntire26srx4, 
title={{    The Fourth Challenge on Image Super-Resolution (×4) at NTIRE 2026: Benchmark Results and Method Overview    }}, 
author={    Chen, Zheng and  Liu, Kai and  Wang, Jingkai and  Yan, Xianglong and  Li, Jianze and  Zhang, Ziqing and  Gong, Jue and  Li, Jiatong and  Sun, Lei and  Liu, Xiaoyang and  Timofte, Radu and  Zhang, Yulun and others    },   
booktitle={Proceedings of the IEEE/CVF Conference on Computer Vision and Pattern Recognition (CVPR) Workshops},  
year = {2026} 
}

@inproceedings{ntire26retouching, 
title={{    Photography Retouching Transfer, NTIRE 2026 Challenge: Report    }}, 
author={    Elezabi, Omar and  V. Conde, Marcos and  Wu, Zongwei and  Jin, Yeying and  Timofte, Radu and others    },   
booktitle={Proceedings of the IEEE/CVF Conference on Computer Vision and Pattern Recognition (CVPR) Workshops},  
year = {2026} 
}

@inproceedings{ntire26rwsr, 
title={{    The First Challenge on Mobile Real-World Image Super-Resolution at NTIRE 2026: Benchmark Results and Method Overview    }}, 
author={    Li, Jiatong and  Chen, Zheng and  Liu, Kai and  Wang, Jingkai and  Zhou, Zihan and  Liu, Xiaoyang and  Zhu, Libo and  Timofte, Radu and  Zhang, Yulun and others    },   
booktitle={Proceedings of the IEEE/CVF Conference on Computer Vision and Pattern Recognition (CVPR) Workshops},  
year = {2026} 
}

@inproceedings{ntire26rsirsr, 
title={{    The First Challenge on Remote Sensing Infrared Image Super-Resolution at NTIRE 2026: Benchmark Results and Method Overview    }}, author={    Liu, Kai and  Yue, Haoyang and  Lin, Zeli and  Chen, Zheng and  Wang, Jingkai and  Gong, Jue and  Timofte, Radu and  Zhang, Yulun and  others    },   
booktitle={Proceedings of the IEEE/CVF Conference on Computer Vision and Pattern Recognition (CVPR) Workshops},  
year = {2026} 
}

@inproceedings{ntire26aigendet, 
title={{    NTIRE 2026 Challenge on Robust AI-Generated Image Detection in the Wild    }}, 
author={    Gushchin, Aleksandr and  Abud, Khaled and  Shumitskaya, Ekaterina and  Filippov, Artem and  Bychkov, Georgii and  Lavrushkin, Sergey and  Erofeev, Mikhail and  Antsiferova, Anastasia and  Chen, Changsheng and  Tan, Shunquan and  Timofte, Radu and  Vatolin, Dmitriy and others    },
booktitle={Proceedings of the IEEE/CVF Conference on Computer Vision and Pattern Recognition (CVPR) Workshops},  
year = {2026} 
}

@inproceedings{ntire26cdfsod, 
title={{    The Second Challenge on Cross-Domain Few-Shot Object Detection at NTIRE 2026: Methods and Results    }}, 
author={    Qiu, Xingyu and  Fu, Yuqian and  Geng, Jiawei and  Ren, Bin and  Pan, Jiancheng and  Wu, Zongwei and  Tang, Hao and  Fu, Yanwei and  Timofte, Radu and  Sebe, Nicu and  Elhoseiny, Mohamed and others    },   
booktitle={Proceedings of the IEEE/CVF Conference on Computer Vision and Pattern Recognition (CVPR) Workshops},  
year = {2026} 
}

@inproceedings{ntire26finrec, 
title={{    NTIRE 2026 Challenge on End-to-End Financial Receipt Restoration and Reasoning from Degraded Images: Datasets, Methods and Results    }}, author={    Guan, Bochen and  Li, Jinlong and  Yang, Kangning and  Ke, Chuang and  Cai, Jie and  Vasluianu, Florin and  Timofte, Radu and others    },   booktitle={Proceedings of the IEEE/CVF Conference on Computer Vision and Pattern Recognition (CVPR) Workshops},  
year = {2026} 
}

@inproceedings{ntire26faceres, 
title={{    The Second Challenge on Real-World Face Restoration at NTIRE 2026: Methods and Results    }}, 
author={    Wang, Jingkai and  Gong, Jue and  Chen, Zheng and  Liu, Kai and  Li, Jiatong and  Zhang, Yulun and  Timofte, Radu and  others    },
booktitle={Proceedings of the IEEE/CVF Conference on Computer Vision and Pattern Recognition (CVPR) Workshops},  
year = {2026} 
}

@inproceedings{ntire26reflection, 
title={{    NTIRE 2026 Challenge on Single Image Reflection Removal in the Wild: Datasets, Results, and Methods    }}, 
author={    Cai, Jie and  Yang, Kangning and  Li, Zhiyuan and  Vasluianu, Florin and  Timofte, Radu and others    },   
booktitle={Proceedings of the IEEE/CVF Conference on Computer Vision and Pattern Recognition (CVPR) Workshops},  
year = {2026} 
}

@inproceedings{ntire26anomalydet, 
title={{    NTIRE 2026  Challenge Report on Anomaly Detection of Face Enhancement for UGC Images    }}, 
author={    Zhong, Yan and   Ma,  Qiufang and  Wang, Zhen and  Jiang, Tingting and  Timofte, Radu and others    },   
booktitle={Proceedings of the IEEE/CVF Conference on Computer Vision and Pattern Recognition (CVPR) Workshops},  
year = {2026} 
}

@inproceedings{ntire26videosal, 
title={{    NTIRE 2026 Challenge on Video Saliency Prediction: Methods and Results    }}, 
author={    Moskalenko, Andrey and  Bryncev, Alexey and  Kosmynin, Ivan and  Shilovskaya, Kira and  Erofeev, Mikhail and  Vatolin, Dmitry and  Timofte, Radu and others    },   
booktitle={Proceedings of the IEEE/CVF Conference on Computer Vision and Pattern Recognition (CVPR) Workshops},  
year = {2026} 
}

@inproceedings{ntire26effsr, 
title={{    The Eleventh NTIRE 2026 Efficient Super-Resolution Challenge Report    }}, 
author={    Ren, Bin and  Guo, Hang and  Shu, Yan and  Ma, Jiaqi and  Cui, Ziteng and  Liu, Shuhong  and  Mei, Guofeng  and  Sun, Lei and  Wu, Zongwei and  Khan, Fahad Shahbaz and  Khan, Salman and  Timofte, Radu and  Li, Yawei and others    },   
booktitle={Proceedings of the IEEE/CVF Conference on Computer Vision and Pattern Recognition (CVPR) Workshops},  
year = {2026} 
}

@inproceedings{ntire26realx3d, 
title={{    3D Restoration and Reconstruction in Adverse Conditions: RealX3D Challenge Results    }}, 
author={    Liu, Shuhong and  Cui, Ziteng and  Bao, Chenyu and  Chu, Xuangeng and  Gu, Lin and  Ren, Bin and  Timofte, Radu and  Conde, Marcos V. and others    },   
booktitle={Proceedings of the IEEE/CVF Conference on Computer Vision and Pattern Recognition (CVPR) Workshops},  
year = {2026} 
}

@inproceedings{ntire26denoising, 
title={{    The Third Challenge on Image Denoising at NTIRE 2026: Methods and Results    }}, 
author={    Sun, Lei and  Guo, Hang and  Ren, Bin and  Su, Shaolin and  Wang, Xian and  Pani Paudel, Danda and  Van Gool, Luc and  Timofte, Radu and  Li, Yawei and others    },   
booktitle={Proceedings of the IEEE/CVF Conference on Computer Vision and Pattern Recognition (CVPR) Workshops},  
year = {2026} 
}

@inproceedings{ntire26aberration, 
title={{    NTIRE 2026 The First Challenge on Blind Computational Aberration Correction: Methods and Results    }}, 
author={    Sun, Lei and  Qian, Xiaolong and  Jiang, Qi and  Wang, Xian and  Gao, Yao and  Yang, Kailun and  Wang, Kaiwei and  Timofte, Radu and  Pani Paudel, Danda and  Van Gool, Luc and others    },   
booktitle={Proceedings of the IEEE/CVF Conference on Computer Vision and Pattern Recognition (CVPR) Workshops},  
year = {2026} 
}

@inproceedings{ntire26eventblurr, 
title={{    The Second Challenge on Event-Based Image Deblurring at NTIRE 2026: Methods and Results    }}, 
author={    Sun, Lei and  Li, Weilun and  Wang, Xian and  Li, Zhendong and  Shi, Letian and  Xu, Dannong and  Zhang, Deheng and  Hu, Mengshun and  Guo, Shuang and  Su, Shaolin and  Timofte, Radu and  Pani Paudel, Danda and  Van Gool, Luc and others    },   
booktitle={Proceedings of the IEEE/CVF Conference on Computer Vision and Pattern Recognition (CVPR) Workshops},  
year = {2026} 
}

@inproceedings{ntire26bursthdr, 
title={{    NTIRE 2026 Challenge on Efficient Burst HDR and Restoration: Datasets, Methods, and Results    }}, 
author={    Park, Hyunhee and  Park, Eunpil and  Lee, Sangmin and  Timofte, Radu and others    },   
booktitle={Proceedings of the IEEE/CVF Conference on Computer Vision and Pattern Recognition (CVPR) Workshops},  
year = {2026} 
}

@inproceedings{ntire26twilight, 
title={{    NTIRE 2026 Low-light Enhancement: Twilight Cowboy Challenge    }}, 
author={    Khalin, Aleksei and  Ershov, Egor and  Panshin, Artem and  Korchagin, Sergey and  Lobarev, Georgiy and  Terekhin, Arseniy and  Dorogova, Sofiia and  Shamsutdinov, Amir and  Mamedov, Yasin and  Khalfin, Bakhtiyar and  Sheludko, Bogdan and  Zilyaev, Emil and  Banić, Nikola and  Perevozchikov, Georgy and  Timofte, Radu and others    },   
booktitle={Proceedings of the IEEE/CVF Conference on Computer Vision and Pattern Recognition (CVPR) Workshops},  
year = {2026} 
}

@inproceedings{ntire26effllie, 
title={{    Efficient Low Light Image Enhancement: NTIRE 2026 Challenge Report    }}, 
author={    Yan, Jiebin  and  Tu, Chenyu  and  Lin, Qinghua and  WU, Zongwei and  Zhang , Weixia and  Wang, Zhihua and  Cao, Peibei and  Fang, Yuming  and  Liu, Xiaoning  and  Zhou, Zhuyun and  Timofte, Radu  and  others    },   
booktitle={Proceedings of the IEEE/CVF Conference on Computer Vision and Pattern Recognition (CVPR) Workshops},  
year = {2026} 
}
}




\end{document}